\documentclass[lettersize,journal]{IEEEtran}
\usepackage{amsmath,amsfonts}
\usepackage{algorithmic}
\usepackage{algorithm}
\usepackage{array}
\usepackage[caption=false,font=normalsize,labelfont=sf,textfont=sf]{subfig}
\usepackage{textcomp}
\usepackage{stfloats}
\usepackage{url}
\usepackage{verbatim}
\usepackage{graphicx}
\usepackage{cite}
\usepackage{amsmath}
\usepackage{amssymb}
\usepackage{booktabs}
\usepackage{multirow}
\usepackage{graphics}
\usepackage{threeparttable}

\newtheorem{condition}{Condition}
\newtheorem{theorem}{Theorem}

\newcommand{\eg}{e.g.}
\newcommand{\ie}{i.e.}

\begin{document}

\title{Towards Vision Transformer Unrolling Fixed-Point Algorithm: a Case Study on Image Restoration}

\author{Peng Qiao, Sidun Liu, Tao Sun, Ke Yang, Yong Dou
\thanks{P.~Qiao, S.~Liu, T.~Sun, Y.~Dou are with the
	Science and Technology on Parallel and Distributed Laboratory,
	School of Computer,
	National University of Defense Technology, Changsha, China.
	e-mail: (pengqiao@nudt.edu.cn).

    K.~Yang is with
    National Innovation Institute of Defense Technology, Beijing, China.
}
}

\markboth{Journal of \LaTeX\ Class Files,~Vol.~14, No.~8, August~2021}%
{Shell \MakeLowercase{\textit{et al.}}: A Sample Article Using IEEEtran.cls for IEEE Journals}

\maketitle

\begin{abstract}
The great success of Deep Neural Networks (DNNs) has inspired the algorithmic development of DNN-based Fixed-Point (DNN-FP) for computer vision tasks. DNN-FP methods, trained by Back-Propagation Through Time or computing the inaccurate inversion of the Jacobian, suffer from inferior representation ability. Motivated by the representation power of the Transformer, we propose a framework to unroll the FP and approximate each unrolled process via Transformer blocks, called FPformer. To reduce the high consumption of memory and computation, we come up with FPRformer by sharing parameters between the successive blocks. We further design a module to adapt Anderson acceleration to FPRformer to enlarge the unrolled iterations and improve the performance, called FPAformer. In order to fully exploit the capability of the Transformer, we apply the proposed model to image restoration, using self-supervised pre-training and supervised fine-tuning. 161 tasks from 4 categories of image restoration problems are used in the pre-training phase. Hereafter, the pre-trained FPformer, FPRformer, and FPAformer are further fine-tuned for the comparison scenarios. Using self-supervised pre-training and supervised fine-tuning, the proposed FPformer, FPRformer, and FPAformer achieve competitive performance with state-of-the-art image restoration methods and better training efficiency. 
FPAformer employs only 29.82\% parameters used in SwinIR models, and provides superior performance after fine-tuning. 
To train these comparison models, it takes only 26.9\% time used for training SwinIR models. It provides a promising way to introduce the Transformer in low-level vision tasks.
\end{abstract}


\begin{figure}[tbp]
	\centering
	\includegraphics[width=0.51\textwidth]{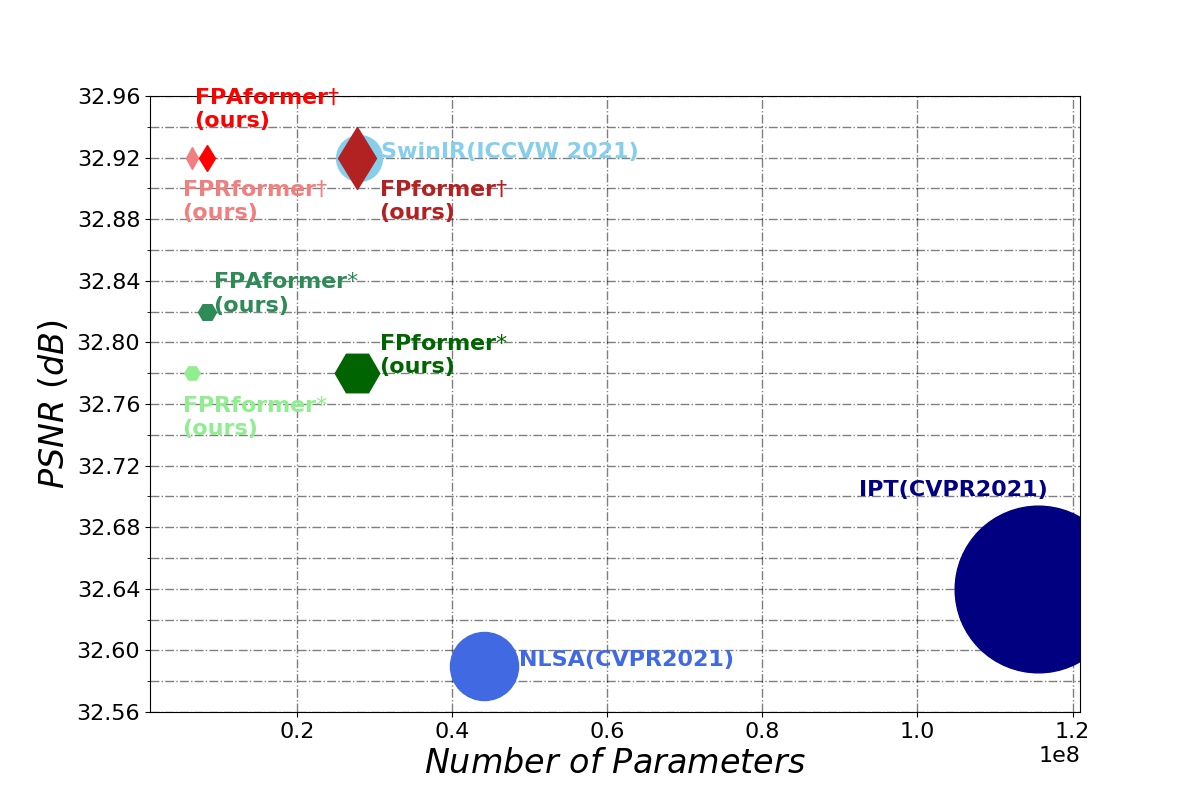}
	\caption{The number of parameters vs SRx4 performance in terms of PSNR on Set5. The \textbf{size of markers} is proportional to the number of parameters in each method. Our proposed methods with $^\dagger$ are finetuned for SRx4, while ones with $^*$ are pretrained.} \label{fig:top}
\end{figure}
\section{Introduction}
\label{sec:intro}

The popular, exuberant, and efficient Deep Neural Networks (DNNs) techniques provide a DNN-based routine for the Fixed-Point (FP) method to handle optimization problems in computer vision \cite{bai2022flow,chang2022obj,fung2022jfb}, decision-making \cite{heaton2021game} and other domains, achieve adorable performance due to existing physical advantages (like GPU computing) or human experiences (like tuning or networks structure settings).

Conventional DNN-FP methods that directly unroll the FP via Convolutional Neural Networks (CNNs) or Recurrent Neural Networks (RNNs) are trained by Back-Propagation (BP) \cite{chen2016tnrd,zhang2017dncnn} or Back-Propagation Through Time (BPTT) \cite{rumelhart1985bptt,werbos1990bptt}. These methods have limited representation ability \cite{chen2016tnrd,qiao2017tnlrd,zhang2017dncnn} and suffer from high memory consumption \cite{gruslys2016memory}, and the gradient vanishing/exploding issues \cite{bengio1994bpttVE,jozefowicz2015empirical,cho2014gru,hochreiter1997lstm} as the number of unrolled depth increases. While Deep EQuilibrium (DEQ) \cite{bai2019deq,bai2020msdeq} methods unroll FP with implicit depth and seek the equilibrium point of FP, they are trained by computing the inversion of the Jacobian of loss w.r.t. the equilibrium point. To alleviate the heavy computation burden of inversion of the Jacobian, an inexact gradient is proposed. However, it yields an undesirable solution.

With the great success of Transformer based models in Natural Language Processing (NLP) \cite{vaswani2017MHSA,lan2019albert,devlin2018bert,brown2020GPT3,shoeybi2019megatron,raffel2020t5} and Computer Vision (CV) \cite{dosovitskiy2021vit,bao2022beit,he2022mae,liu2021swin,liang2021swinir,chen2021ipt,li2021edt,zamir2022restormer}, it has been shown that Transformer-based models are suitable for model the sequential relation with a powerful representation. Motivated by this fact, we propose to unroll the FP and approximate each unrolled process via Transformer called FPformer. 
Nevertheless, Transformer-based methods increase the consumption of memory and computation.

To handle this issue, we investigate the parameter sharing \cite{lan2019albert} in FPformer, called FPRformer. In this setting, the successive blocks in Transformers share the parameters, resulting in fewer trainable parameters and maintaining the unrolled iteration times. 

Based on our analysis in Section \ref{sec:method_fpaformer}, we further apply Anderson acceleration \cite{anderson1965aa,toth2015aac} to FPRformer via a simplified ConvGRU \cite{cho2014gru,teed2020raft} module to enlarge the unrolled iterations, called FPAformer.

To verify the effectiveness of the proposed FPformer and its variants, we apply it to image restoration task sets as a general image restoration framework. In order to fully exploit the capability of the Transformer, we train the FPformer, FPRformer, and FPAformer using self-supervised pre-training and supervised fine-tuning, widely used in NLP and high-level vision tasks. In the self-supervised pre-training, fixed-point finding for solving image restoration problems becomes a natural interpretation for general image restoration, serving as the self-supervised pre-training problem. We use 161 tasks from 4 categories of image restoration problems to pre-train the proposed FPformer, FPRformer, and FPAformer. Namely, the image restoration tasks are Gaussian denoising in grayscale and color space (noise levels ranging from 0 to 75), single image super-resolution (scale factors are 2, 3, 4, and 8), and JPEG deblocking (quality factors are 10, 20, 30, 40 and 50). During the supervised fine-tuning, the pre-trained FPformer, FPRformer, and FPAformer are further fine-tuned for a specific comparison scenario, \eg, Gaussian denoising in color space with noise level $\sigma=25$.
Using self-supervised pre-training and supervised fine-tuning, the proposed FPformer, FPRformer, and FPAformer achieve competitive performance with state-of-the-art image restoration methods and better efficiency, as shown in Figure \ref{fig:top}, providing a promising way to introduce Transformer in low-level vision tasks.

\begin{figure*}[ht]
	\centering
	\includegraphics[width=0.8\textwidth]{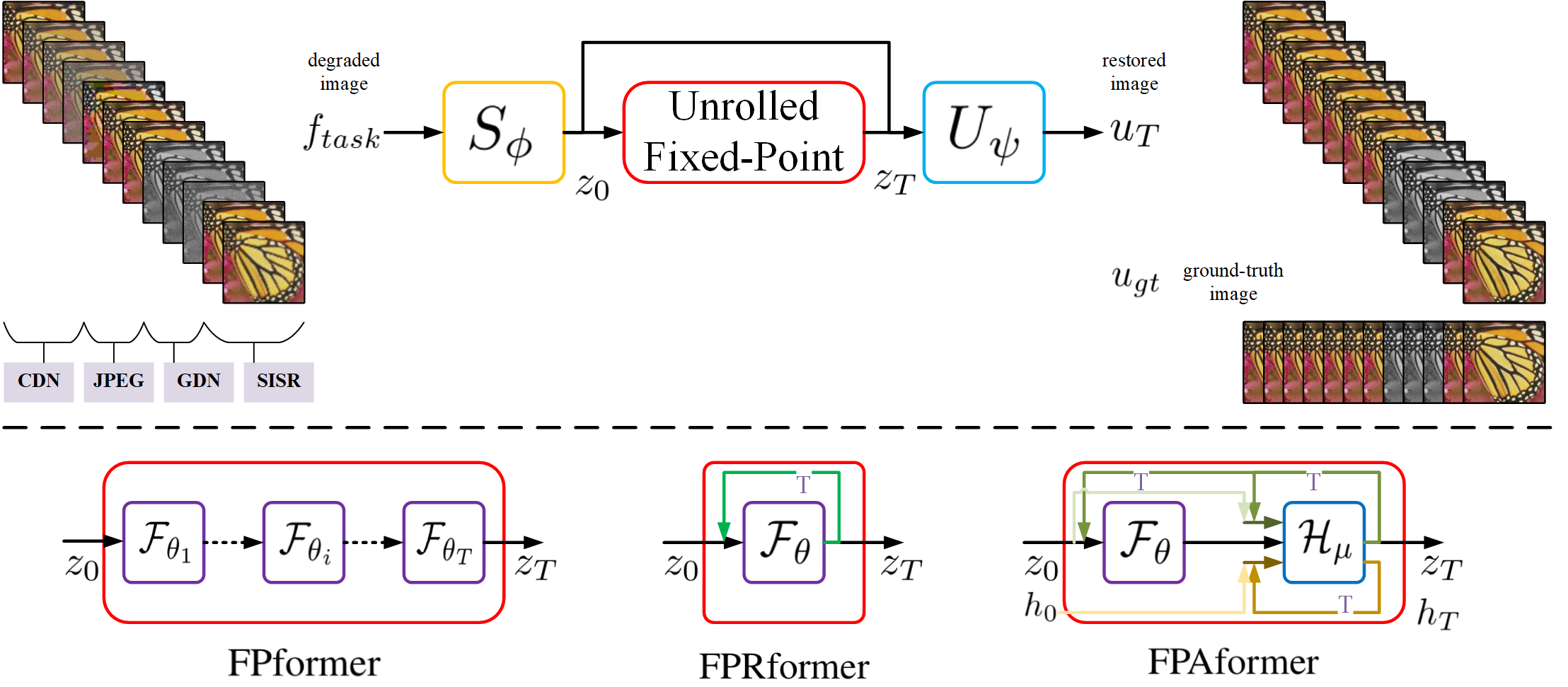}
	\caption{The architecture of the proposed FPformer, FPRformer, and FPAformer.\textbf{CDN} and \textbf{GDN} mean Color Gaussian Denoising and Grayscale Gaussian Denoising, respectively. \textbf{JPEG} stands for image JPEG deblocking. \textbf{SISR} means Single Image Super-Resolution.}
	\label{fig:arch}
\end{figure*}

\section{Related works}
\label{sec:related}
\subsection{Fixed-Point via DNNs}
The fixed-point is formulated as
\begin{equation}\label{eq:fp}
	z^* = \mathcal{F}(z^*).
\end{equation}
The fixed-point finding in Algorithm \ref{alg:fpf}, details in supplementary materials, generates a series of $\{{z_t}\}_{t=1}^{T}$ by successively applying the contraction mapping $\mathcal{F}(.)$, given a initial point $z_0$. 
When we focus on the states $z_t$, we can simplify the Algorithm \ref{alg:fpf} as
\begin{equation}\label{eq:unroll1}
	z_0 \stackrel{\mathcal{F}(.)}{\longrightarrow} z_1 \longrightarrow \cdots \longrightarrow z_{T-1} \stackrel{\mathcal{F}(.)}\longrightarrow z_T.
\end{equation}

Conventional DNN-FP methods that directly unroll the FP via Convolutional Neural Networks (CNNs) or Recurrent Neural Networks (RNNs), \ie, parameterizing $\mathcal{F}(.)$ with $\mathcal{F}_{\theta}$. These methods either have limited representation ability \cite{chen2016tnrd,qiao2017tnlrd,zhang2017dncnn} or suffer from high memory consumption \cite{gruslys2016memory} and the gradient vanishing/exploding issues \cite{bengio1994bpttVE,jozefowicz2015empirical,cho2014gru,hochreiter1997lstm} as the number of unrolled iterations increases. While Deep EQuilibrium (DEQ) \cite{bai2019deq} methods unroll the FP with implicit depth and seek the equilibrium point of the FP, they are trained by computing the inversion of the Jacobian of loss w.r.t. the equilibrium point. DEQ and its variants \cite{fung2022jfb} suffer from high computation and inexact gradient to achieve an unacceptable solution. In \cite{gilton2021deq}, DEQ is applied in solving inverse problems in imaging, where $\mathcal{F}(.)$ is a specific proximal operator and is further parameterized via $\theta$.

In the fixed-point finding method, the iteration process in Algorithm \ref{alg:fpf} requires quite a lot of iterations, \ie, a large $T$, to reach a feasible equilibrium point $z^*$. One should also note that the choice of hyper-parameters of $\epsilon$ and $T$ is vitally important to achieving a good performance using $\mathcal{F}_{\theta}$, whose contraction property is not well-guaranteed. When applying DNN-FP methods, repeating the modern DNN a couple of times is computation-consuming and time-consuming. For example, DEQ and its variants still consume GPU memory as large as modern DNN, and even more computational consumption to perform the fixed-point finding algorithm to achieve a reasonable performance. 

Anderson acceleration (\cite{anderson1965aa,toth2015aac}, AA) is proposed to accelerate the fixed-point finding utilizing the previous $m$ states $\{z_{t-m+i}\}_{i=1}^{m}$ to estimate the next state $z_{t+1}$, as shown in Algorithm \ref{alg:fpaa} in supplementary materials.
In \cite{bai2021ndeq}, Anderson acceleration is integrated into DEQ and parameterized via NNs. In order to explicitly combine the preview $m$ states, the proposed AA module exploits a bottle-neck-like architecture to produces real value weights. Thus, the preview $m$ states are needed to be buffered.
To save the storage costs, in \cite{bai2021ndeq}, the bottle-neck-like network that benefits NLP but degrades image processing because images are represented in 2d-data and have vivid context information in spatial domains while natural languages only need 1d-data. 

\subsection{Image Restoration}
Conventional image restoration methods can be used to recover various image degradation problems by minimizing the following energy function, 
\begin{equation}\label{eq:ebm}
	\mathcal{E}(u,f)=\mathcal{D}(u,f)+\lambda \mathcal{R}(u),
\end{equation}
where $\mathcal{D}(u,f)$ is the data term related to one specific image restoration problem, $f$ is the degraded input image, and $u$ is the restored image. Taking Gaussian denoising as an example, $\mathcal{D}(u,f)=\frac{1}{2{\sigma}^2} \| u -f\| ^2$, where $\sigma$ is the noise level for a specific Gaussian denoising problem. $\mathcal{R}(u)$ is the regularization term known as the image prior model \cite{rudin1992tv,buades2005nlmeans,dabov2007bm3d,danielyan2011bm3d,gu2014wnnm,chen2016tnrd}. Empirically, one can get a minimizer of Equation \ref{eq:ebm} via gradient descent. It can be reformulated into a fixed-point finding diagram when $\mathcal{I}-\nabla \mathcal{E}$ is a contraction mapping.

Benefiting from machine learning methods, the above hand-crafted methods can be further boosted. For example, the diffusion models in \cite{chen2016tnrd,qiao2017tnlrd} are the learned counterparts of their conventional ones. With the rapid development of CNNs and Transformers, the restoration methods are learned in a data-driven manner and provide very impressive performance in image denoising \cite{zhang2021drunet,zhang2017dncnn}, single image super-resolution \cite{wang2018esrgan,zamir2021MPRNet}, JPEG deblocking \cite{zhang2020rdn,zhang2021drunet,fu2021dejpeg}, image deblurring \cite{li2018bid,tao2018srd}, et. al.

Most of CNNs based image restoration methods can be regarded as learning the mapping between the degraded input image $f$ and its corresponding ground-truth image $u_{gt}$. Therefore, we can summarize that these CNNs based methods are parameterized in the very first iteration in Equation \ref{eq:unroll1}, \ie, $z_1=\mathcal{F}_{\theta}(z_0)$. $z_0$ is mapped from the degraded image $f$ via $S_{\phi}$. $z_T \big|_{T=1}$ is mapped to restored image via $U_{\psi}$. Note that, both $S_{\phi}$ and $U_{\psi}$ are identical mapping in \cite{zhang2017dncnn}, and convolution operators in \cite{liang2021swinir}. Due to large parameters and low training efficiency, approximating each step $z_t$ using CNNs is not a common practice. On the contrary, TNRD \cite{chen2016tnrd} and its variants\cite{qiao2017tnlrd} are trained to approximate the unrolling of Equation \ref{eq:unroll1} in fixed iterations. In each iteration, the contraction mapping $\mathcal{F}_{\theta_t}$ is a learned diffusion model parameterized by $\theta_t$. However, the representation ability of TNRD is quite limited. So, we need an effective way to approximate each step $z_{t}$ in Algorithm \ref{alg:fpf}.

\subsection{Vision Transformer}
Multi-Head Self Attention (MHSA) introduced in \cite{vaswani2017MHSA} has been widely used in natural language processing \cite{devlin2018bert,brown2020GPT3,shoeybi2019megatron,raffel2020t5}. In \cite{dosovitskiy2021vit}, MHSA was adopted in computer vision, resulting in vision transformers. And since then, vision transformers \cite{dosovitskiy2021vit,bao2022beit,liu2021swin,he2022mae} trained by self-supervised pre-training and supervised fine-tuning have achieved better performance than CNN models in high-level vision.

In low-level vision represented by image restoration problems, vision transformer architecture is adapted \cite{liang2021swinir,chen2021ipt,li2021edt,zamir2022restormer,tu2022maxim} and also benefits the performance compared with the CNN counterparts \cite{zhang2017dncnn,zhang2021drunet}. These works trained transformer models via supervised training for one specific task and even one specific task setting. In \cite{liang2021swinir,zamir2022restormer}, the proposed methods were trained for Gaussian denoising with different noise levels, for single image super-resolution with different scales, respectively. In \cite{chen2021ipt}, IPT was trained for a couple of image restoration tasks with an auxiliary token, \ie, task-specific token. In \cite{zamir2022restormer}, Restormer is proposed and provides a way to efficiently train for high-resolution image restoration problems. Therefore, the self-supervised pre-training for image restoration is worth discussing to achieve an efficient way to utilize a vision transformer.

\section{Methodology}\label{sec:method}
In this section, we first describe the proposed FPformer approximating the unrolling of the FP. To reduce the memory consumption, we utilize parameter sharing between the successive blocks in FPformer, called FPRformer. To boost the restoration performance, we design a module inspired by the Anderson acceleration algorithm, called FPAformer.

\subsection{Unrolling FP with Transformers: FPformer}\label{sec:method_fpformer}

As described above, DEQ and its variants benefit from the implicitly infinite iterations to conduct fixed-point finding with a large $T$ and small $\epsilon$; they need a huge computational consumption to reach the FP when $\mathcal{F}$ is approximated via modern DNNs. Directly learning the unrolling of the fixed-point finding in fixed iterations, \eg, DnCNN with $T=1$ and TNRD with a larger $T$ less than 10, it is limited by the unrolled times $T$ or the representation ability $\mathcal{F}_{\theta_t}$. We need an effective way to enlarge the unrolled times and strengthen the representation ability.

Having witnessed the success of NLP and CV, the Transformer-based models are suitable for modeling the sequential relation with a powerful representation. Motivated by this fact, we propose to unroll the FP and approximate each unrolled process via Transformer called FPformer.
To this end, we resort to Transformers. To efficiently capture global information, we use Residual Swin Transformer Block (RSTB) blocks as proposed in \cite{liang2021swinir} to learn each contraction mapping in Equation \ref{eq:unroll1}, $\mathcal{F}_{\theta_t}, t=1,\cdots,T$. 
\begin{equation}\label{eq:unroll2}
	z_0 \stackrel{F_{\theta_1}}{\longrightarrow} z_1 \longrightarrow \cdots \longrightarrow z_{T-1} \stackrel{F_{\theta_{T}}}\longrightarrow z_T.
\end{equation}

The resulting architecture for image restoration is called FPformer and is shown in Figure~\ref{fig:arch} and Equation \ref{eq:unroll2}. 

Naturally, the fixed-point finding in Equation \ref{eq:unroll1} of minimizing Equation \ref{eq:ebm} is agnostic to the image restoration task. It just tries to recover a degraded image $f$ to a restored image $u_T$, which is close to the ground-truth clean image $u_{gt}$. Therefore, to fully explore the capability of the Transformer, we trained the FPformer using multiple image restoration problems. 

The training of FPformer is formulated as 
\begin{align}\label{eq:train_fpformer}
	\begin{aligned} 
		& \mathop{\min}_{\Theta} \sum_{s \in S} \mathcal{L} (u_{gt}^s, u_T^s) \\
		z_0^s &= S_{\phi}(f_{task}^s) \\
		z_T^s & =\mathcal{F}_{\theta_{T}}(\cdots (\mathcal{F}_{\theta_1}(z_0^s)))= \text{FPformer}_{\Theta}(z_0^s)\\
		u_T^s &= U_{\psi}(z_T^s)
	\end{aligned}
\end{align}
where $\Theta=\{\theta_1,\cdots,\theta_{T}\}$ is the parameters in FPformer, $\mathcal{L}$ is the loss function that measures the difference between the ground-truth image $u_{gt}^s$ and the restored image $u_T^s$. $u_{gt}^s, f_{task}^s$ are generated using a specific sample in dataset $S$, details are described in Section \ref{sec:exp}. Following \cite{liang2021swinir}, both $S_{\phi}$ and $U_{\psi}$ are convolution operators. To be specific, $S_{\phi}$ represents the degraded image $f_{task}\in \mathbb{R}^{B\times H \times W \times 3}$ as the first state $z_0 \in \mathbb{R}^{B\times H \times W \times C}$. $B$ is the size of the minibatch, $H$ and $W$ are the height and width of the image (or patch), and $C$ is the channel number of features. $U_{\psi}$ maps the last state $z_T \in \mathbb{R}^{B\times H \times W \times C}$ to the restored image $u_T\in \mathbb{R}^{B\times H \times W \times 3}$. 

In FPformer, both $S_{\phi}$ and $U_{\psi}$ are shared among image restoration problems and specific tasking settings. To handle different upscales in single image super-resolution, instead of upscaling the features via upsampling blocks, we upscaled the downscaled images with the corresponding scaling factor when preparing the minibatch. Details are discussed in Section \ref{sec:exp}. In SwinIR\cite{liang2021swinir}, $S_{\phi}$ and $U_{\psi}$ are task-specific convolutional operators. In single image super-resolution, dedicated upsampling modules are used. In IPT \cite{chen2021ipt}, $S_{\phi}$ and $U_{\psi}$ are also task-specific, only Transformer blocks are shared. In Restormer \cite{zamir2022restormer}, the degraded and restored image have the same resolution. It may explain why single image super-resolution is not discussed. In summary, FPformer can be treated as a general image restoration solver. Detail performance is discussed in Section \ref{sec:exp}. 

\subsection{Sharing parameters: FPRformer}\label{sec:method_fprformer}
In ALBERT \cite{lan2019albert}, sharing parameters among Transformer blocks results in fewer amounts of models parameters and smaller model sizes. In this context, one can maintain the depth of the Transformers while regularizing the whole Transformers, resulting in a small model while providing a competitive performance. 

Inspired by this idea, we enforce successive $N_j$ blocks in the $T$ blocks of FPformer to share parameters, coined as FPRformer. Therefore, we have $\sum_{j=1}^{R} N_j = T$, where $R$ is the number of unique RSTB blocks; $N_j$ is the recurrent times of $j_{th}$ unique RSTB block. The amounts of the parameters are about $R/T$ times less compared with FPformer with $T$ RSTB blocks. Note that FPformer can be regarded as a special case of FPRformer with $R=T$ and $N_j = 1, \forall j \in \{1, \cdots, R\}$.

We trained FPRformer with $R=2$, $R=3$ and $R=T$, separately. The details of this ablation study are shown in Section \ref{sec:exp_abla}.

\subsection{Anderson Acceleration: FPAformer}\label{sec:method_fpaformer}
We present a theorem to characterize the performance of \eqref{eq:unroll2}, whose details can be found in supplementary materials.
\begin{theorem}\label{th0}[Informal]
	If the trained model $(\theta_t)_{1\leq t\leq T}$ can fit $\mathcal{F}$ well. Let $z^*$ be the fixed point of $\mathcal{F}$, we have
	\begin{equation} 
		\|z_T-z^*\|=\mathcal{O}\Big(\rho^T+\frac{\delta}{1-\rho}\Big)
	\end{equation}
	for some fixed $0<\rho<1$ and $\delta\geq 0$ reflecting how the model fits (a smaller $\delta$ indicate better fitting).
\end{theorem}
Based on Theorem \ref{th0}, we can immediately get two claims.
\begin{itemize}
	\item As $T$ increases, the bound of $\|z_T-z^*\|$ gets small. That indicates when we use a larger $T$, the unrolling yields better results.
	\item When $T=\frac{\ln\frac{1-\rho}{\delta}}{\ln\frac{1}{\rho}}$, it holds $\|z_T-z^*\|=\mathcal{O}(\frac{\delta}{1-\rho})$. That means as $T$ is fixed as some integer, the performance is only then determined by how the model fits.
\end{itemize}

Following the above two claims, we conclude that the performance of FPRformer may lag behind that of FPformer because of its parameter-sharing setting. To boost the performance of FPRformer while enjoying the parameter sharing, we design a module analog to the Anderson acceleration to explicitly enlarge the iteration times and get $\delta$ smaller.

As described in Algorithm \ref{alg:fpaa}, the Anderson acceleration algorithm accelerates the FP using the previous states. It is complex to directly translate the Algorithm \ref{alg:fpaa}, especially lines 5-7, into CNNs or RNNs. It is because the forward and backward computation involves the previous outcome of FP and causes nested dependency. This will take more GPU memory and cause more complicated computational graphs which harms the GPU performance greatly. To this end, we simplify the computation in Algorithm \ref{alg:fpaa} into a recurrent module depending on the current state $z_t$ and hidden state $h_t$. Let hidden state $h_t$ to maintain and summarize the previous $m$ states, $\{\mathcal{G}_{t-m_t+1}, \cdots, \mathcal{G}_{t}\}$. In this way, the simplified Anderson acceleration of the fixed-point finding algorithm is formulated as
\begin{align}\label{eq:aafunc}
	\begin{aligned}
		\hat{z}_{t+1} &= \mathcal{F}_{\theta} (z_t), \\
		z_{t+1}, h_{t+1} &=\mathcal{H}_{\mu}(z_t, \hat{z}_{t+1}, h_t),
	\end{aligned}
\end{align}
where $\mathcal{H}_{\mu}$ is parameterized by $\mu$. Line 6 of Anderson acceleration algorithm \ref{alg:fpaa} determines the weights of the previous states. Then the weights combine these states to update the next state, as shown in Line 7 of \ref{alg:fpaa}. In our simplified version (\ref{eq:aafunc}), the weights calculation and combination can be summarized as GRU \cite{cho2014gru}. Follow \cite{teed2020raft}, we adopt ConvGRU module to learning  $\mathcal{H}_{\mu}$. The proposed simplified ConvGRU module is as follows,
\begin{align}\label{eq:convgru}
	\small
	\begin{aligned}
		\mathcal{G}_t&= \mathbf{Conv}(\hat{z}_{t+1} -z_t ), \\
		r_h&= \sigma(\mathbf{Conv}(\mathcal{G}_t)+\mathbf{Conv}(h_t )), \\
		r_z&= \sigma(\mathbf{Conv}(r_h )), \\
		h_{t+1}&= \mathbf{Norm}((1-r_h ) \odot h_t+r_h \odot \mathbf{Conv}(\mathcal{G}_t)), \\
		z_{t+1}&= \mathbf{Norm}((1-r_z ) \odot \hat{z}_{t+1} + r_z \odot \mathbf{Conv}(\mathcal{G}_t)),
	\end{aligned}
\end{align}
To analog the $m$ previous states setting in Algorithm \ref{alg:fpaa}, the hidden states $h_t\in \mathbb{R}^{B\times H \times W \times mC}$ is $m$ times larger than $z_t\in \mathbb{R}^{B\times H \times W \times C}$. $\mathbf{Conv}$ is 2d convolutional layer. $\sigma(.)$ is the sigmoid function. $\mathbf{Norm}$ is the layer normalization \cite{ba2016ln}. Layer normalization is commonly used in Transformers and benefits the convergence of the training. Therefore, we add $\mathbf{Norm}$ on the output of $h_{t+1}$ and $x_{t+1}$. $\odot$ is the element-wise product. $h_0$ is initialized as all zero values.

\section{Experiments}\label{sec:exp}
\subsection{Experimental Setup}
\noindent\textbf{Image restoration tasks in training.}
We use commonly used image restoration tasks, \eg, color and gray Gaussian denoising, single image super-resolution (SISR), and image JPEG deblocking in training. \textbf{For color and gray Gaussian denoising}, we obtain noisy images by adding additive white Gaussian noises with noise level $\sigma$ ranging from 0 to 75. \textbf{For SISR}, we downscale and upscale\footnote{We use code from \url{https://github.com/fatheral/matlab_imresize/}.} images with scale 2, 3, 4, 8. Note that instead of upscaling the features $z_t$ via upsampling blocks in FPformer, we upscaled the downscaled images $f$ with the corresponding scaling factor. So, the resulting images $u_T$ are in the same spatial resolution as the ground-truth images $u_{gt}$. \textbf{For image JPEG deblocking}, we generate low-quality images using a JPEG encoder with quality factor $q$=10, 20, 30, 40, 50. Therefore, we train FPformer for 161 tasks from 4 categories of image restoration problems simultaneously. FPRformer and FPAformer are trained in the same setting. 

\noindent\textbf{Training datasets.}\label{sec:exp_dataset}
Following \cite{liang2021swinir,wang2018esrgan}, we train FPformer, FPRformer and FPAformer in the above image restoration tasks using random cropped patches from 800 images in DIV2K \cite{agustsson2017div2k}, 2560 images in Flickr2K \cite{timofte2017flickr2k}, 300 images in BSD500 \cite{arbelaez2010bsd500} and all images in WED \cite{ma2016wed}. For DIV2K, we use the first 800 images. For Flickr2K, we use the first 2560 images. For BSD500, we have the whole 300 images in the trainset. For WED, we use the whole dataset.

\noindent\textbf{Pre-Training.}
All training of FPformer, FPRformer, and FPAformer is run on a server with 8 NVIDIA GeForce V100 GPUs. The batch size is 16. The patch sizes are $48 \times 48$, $72 \times 72$, $120 \times 120$ (window size is $8 \times 8$). The RSTB block is set as follows. In FPformer, the number of RSTB blocks is 9 ($T=9$). In FPRformer and FPAformer, the influences of $T$ are discussed in Section \ref{sec:exp_abla}. In each RSTB block, the number of Swin Transformer Layer is 6; the channel number $C$ is 240. The head number of MHSA is 8. 

When preparing the training minibatch, we first sample clean images from the above training datasets, crop them into patches (with the above patch size, \eg $48 \times 48$), and augment these patches. Then we randomly choose an image restoration problem from color and gray Gaussian denoising, SISR, and image JPEG deblocking for each augmented patch. For the chosen problem, we randomly choose the task setting, \ie, the noise level $\sigma$, the scale, or the quality factor $q$. Then we apply the chosen image degradation and setting to each augmented patch. The resulting degraded patches and the augmented clean patches consist of training pairs. 

Note that the degraded patches in training pairs are generated on-the-fly instead of generating degraded images offline as \cite{liang2021swinir}. We augment the training images using color space convert augmentation, flipping, rotating and other data augmentation methods as \cite{liang2021swinir}. 

The learning rate is halved at [500K, 750K, 900K, 950K, 1000K], the initial learning rate is 2e-4. We use Adam \cite{kingma2014adam} optimizer with $\beta_1 = 0.9$ and $\beta_2 = 0.99$. The training loss $\mathcal{L} (u_{gt}, z_{T} )$ in Equation \ref{eq:train_fpformer} is the Charbonnier loss \cite{charbonnier1994Charbonnier}.

\noindent\textbf{Fine-Tuning.} 
When the above pre-training of FPformer, FPRformer, and FPAformer is done, each of them is capable to restore the degraded images from those image restoration tasks. To boost the performance for a specific image restoration problem or task setting, we further finetuned FPformer, FPRformer, and FPAformer. The initial learning rate and schedule are discussed in Section \ref{sec:exp_abla}.

\noindent\textbf{Evaluation.} We pad the image in testing so that the image size is a multiple of the window size. We compare the proposed FPformer, FPRformer, and FPAformer with the previous state-of-the-art methods in color and gray Gaussian denoising, SISR, and image JPEG deblocking. The performance metrics are PSNR and SSIM \cite{liang2021swinir}. The evaluation details are shown in \ref{sec:exp_sota}. 

\begin{figure*}[tbph!]
	\centering
	\includegraphics[width=0.9\textwidth]{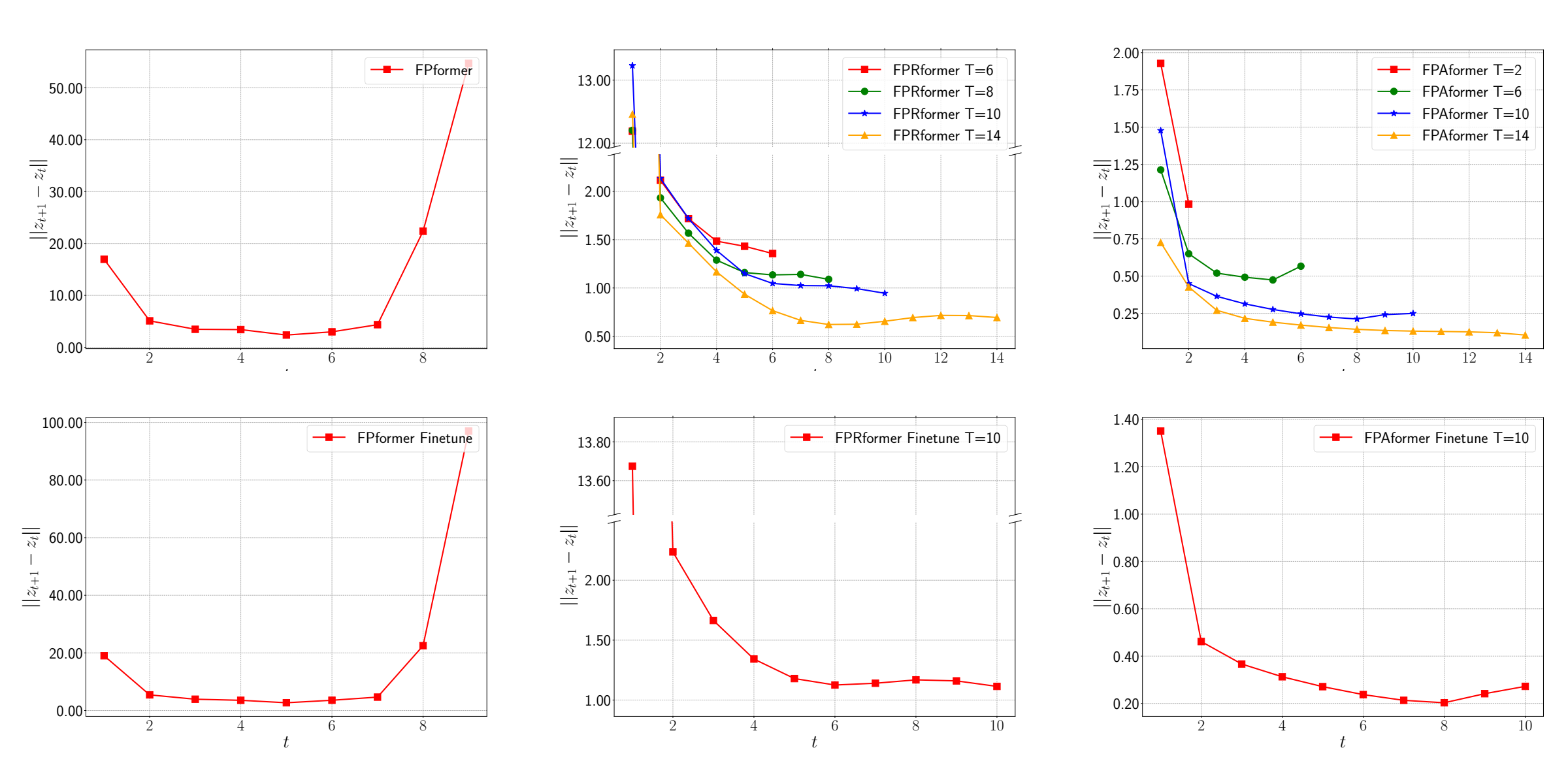}
	\caption{The $l_2$ distances of input $z_t$ and output $z_{t+1}$ for each layer in SRx4. From left to right, each column FPformer, FPRformer, and FPAformer. The first row is for pre-trained FPformer and its variants. The second row is for fine-tuned models for SRx4.} \label{fig:abla_dist}
\end{figure*}
\subsection{Ablation Studies}\label{sec:exp_abla}
In ablation studies, we investigate the influence of key hyper-parameters on the performance of FPformer, FPRformer, and FPAformer.  
The RSTB block is set as above. We compare the performance on Set5 \cite{bevilacqua2012set5} in SISR with scale 4 (SRx4).

\noindent\textbf{The recurrent times $T$ and patch size.}
We trained FPRformer with $R=2$ and $N_1=1$, $N_2=T-1$ given $T$ (6, 8, 10, and 14). We trained FPformer with the default RSTB setting for the different patch sizes listed above. The result is summarized in Table \ref{tab:abla_fprformer}. 
 
The number of parameters in FPformer with different patch sizes is the same, about 27.8M. The number of parameters in FPRformer is 6.5M. Increasing the patch size in training is able to extend the image context, which in turn benefits the performance of FPformer and its variants. Therefore, we believe the performance of FPformer and its variants will continue to increase with larger patch size, as progressive training in \cite{zamir2022restormer}. Due to the limited GPU memory, we were not able to train FPformer and its variants with patch size larger than $120 \times 120$. 

\begin{table}[bt!]
	\caption{The influence of $T$ and patch size. The evaluation metric is PSNR.}
	\label{tab:abla_fprformer}
	\begin{center}
		{\begin{tabular}{c|c|c|c|c|c}
				\hline
				patch  &  \multirow{2}*{FPformer} & \multicolumn{4}{c}{FPRformer} \\
				\cline{3-6} 
				size &  & $T=6$ & $T=8$ & $T=10$ & $T=14$ \\
				\hline
				48 & 32.52 & 32.37 & 32.46 & 32.46 & 32.49 \\
				72 & 32.68 & 32.54 & 32.62 & 32.64 & 32.61 \\
				120 & 32.78 & 32.64 & 32.76 & 32.78 & 32.78 \\
				\hline
		\end{tabular}}
	\end{center}
\end{table}

As the increase of $T$, the performance of FPformer, FPRformer, and FPAformer increases, as shown in Table \ref{tab:abla_fprformer} and Table \ref{tab:abla_fpaformer}. The performance of FPAformer improves as $T$ increases, even achieving better performance than FPformer. 
One can find that the performance of FPRformer and FPAformer is on-par with FPformer. The parameter number of FPRformer and FPAformer is reduced to $23.4\%$ and $30.2\%$ of FPformer's, respectively. 

\noindent\textbf{The number of previous states $m$ in $\mathcal{H}_{\mu}$.}
We trained FPAformer with different kernel sizes, $ks$, and various numbers of previous states $m$. The result is summarized in the upper table of \ref{tab:abla_fpaformer}. When $ks=1$, as the increase of $m$ the performance peak is presented in $m=3$. $m$ is set to analog the number of previous states used in Anderson acceleration. Too large or too small $m$ harms the performance. When $m=3$, as the increase of $ks$, the performance is degraded. This indicates that spatial fusion via a larger convolutional kernel is not as important as temporal fusion via larger $m$.

\begin{table}[tpb]
	\caption{For a given patch size 48 and 120, $R=2$. In the first super block, $T=10$, the influence of $m$ and kernel size (ks) in $\mathcal{H}_{\mu}$. In the second super block, $ks=1$, $m=3$, the influence of $T$ in FPAformer. The evaluation metric is PSNR.}
	\label{tab:abla_fpaformer}
	\begin{center}
		{\begin{tabular}{c|c|c|c|c|c}
				\hline
				patch & \multirow{3}*{FPformer} & \multicolumn{4}{c}{FPAformer ($T=10$)} \\
				\cline{3-6} 
				size & & $ks=1$ & $ks=1$ & $ks=1$ & $ks=3$ \\
				& & $m=1$  & $m=3$  & $m=5$  & $m=3$  \\
				\hline
				48  & 32.52 & 32.47 & 32.52 & 32.48 & 32.45 \\
				120 & 32.78 & -     & 32.82 & -     & -  \\
				\hline
		\end{tabular}}
		{\begin{tabular}{c|c|c|c|c|c}
				\hline
				patch & \multirow{3}*{FPRformer} & \multicolumn{4}{c}{FPAformer} \\
				size  & & \multicolumn{4}{c}{$ks=1$,$m=3$} \\
				\cline{3-6} 
				& & $T=2$ & $T=6$ & $T=10$ & $T=14$ \\
				\hline
				48  & 32.46 & 32.06 & 32.47 & 32.52 & 32.52 \\
				120 & 32.78 & -     & 32.70 & 32.82 & -  \\
				\hline
		\end{tabular}}
	\end{center}
\end{table}

\noindent\textbf{Contraction mapping.}
Parameter sharing in FPRformer and FPAformer benefits the memory footprint of the models. It is natural to ask why FPRformer and FPAformer exploit fewer parameters while providing competitive performance. And what else can we learn from this setting?

As shown in Figure \ref{fig:abla_dist}, the $l_2$ distances of input $z_t$ and output $z_{t+1}$ for each layer of FPformer becomes larger. On the contrary, those of FPRformer and FPAformer are narrowing down but do not get close to zero. Meanwhile, the minimum $l_2$ distance of FPAformer is smaller than FPRformer. Intuitively, both FPRformer and FPAformer seem to be contractive. It may indicate that FPRformer and FPAformer behave like seeking the fixed-point of image restoration problems. More curves in Gaussian denoising and image JPEG deblocking are shown in the supplementary materials.

\begin{figure}[tbp!]
	\centering
	\includegraphics[width=0.4\textwidth]{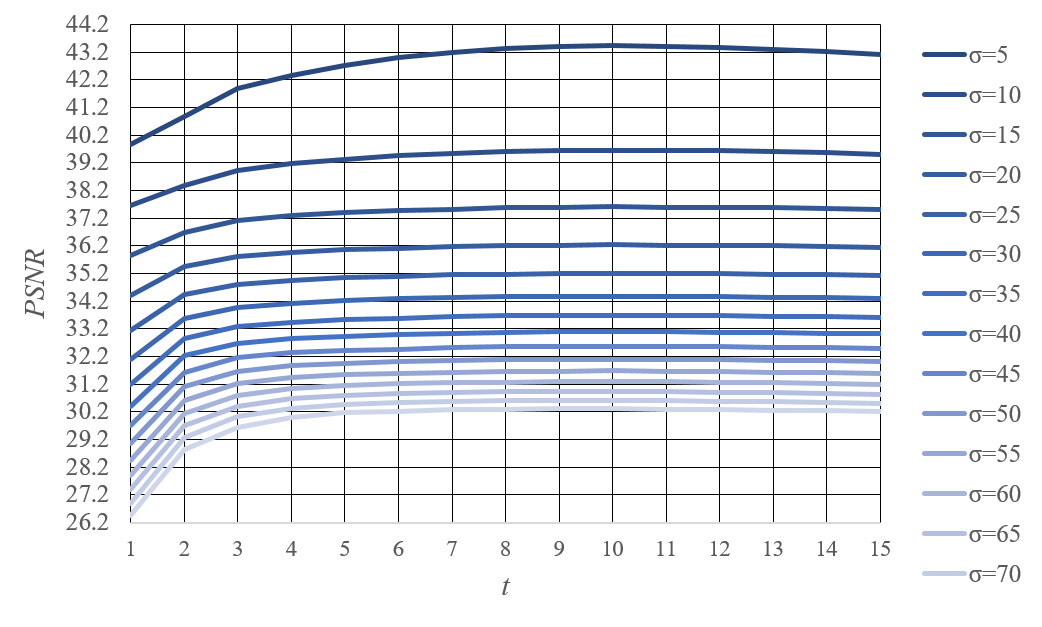}
	\caption{The behavior when repeating FPRformer with another iterations $T=15$, larger than their training setting, $T=10$.} \label{fig:abla_multiout}
\end{figure}

We further investigate the behavior when repeating FPRformer with another iteration $T=15$, larger than their training setting, $T=10$. The result is shown in Figure \ref{fig:abla_multiout}. We ran FPRformer in color Gaussian denoising ranging from 5 to 75 with step 5. These noises were added to images in Set5. The denoising performance for each $\sigma$ was averaged over images in Set5. Figure \ref{fig:abla_multiout} indicates that the peak performance is achieved around $T=10 \pm 2$. As iteration exceeds $T=10$, the performance almost stays unchanged. Considering both phenomena, in FPRformer and FPAformer, one can shorten $T$ to balance between the image restoration performance and the inference speed.

The ablation studies reveal that the ability to provide almost the same performance while sharing parameters among blocks comes from the increase of $T$. As $T$ gets larger, the distance in the final $T_{th}$ layer gets smaller, the performance gets better. It seems that FPRformer is seeking the contraction point in the feature space. As mentioned in \cite{bai2019deq}, deep equilibrium models are implicitly seeking the equilibrium point of Equation \ref{eq:unroll1}. On the contrary, FPformer, FPRformer, and FPAformer are trained to find the equilibrium point of Equation \ref{eq:unroll1} with $T$ steps, each of them learned via Transformer blocks. Meanwhile, our proposed methods avoid the computation or approximation of the inverse Jacobian in \cite{bai2019deq}.

\begin{table*}[hbt!]
	\caption{Comparison with state-of-the-art methods for single image super-resolution on benchmark datasets in terms of average PSNR/SSIM. The best and second-best results are highlighted and underlined, respectively. $\dagger$ means Fine-tuned for a specific task. $*$ means Pre-trained.}
	\label{tab:sr}
	\begin{center}
		\begin{threeparttable}
			{
				\begin{tabular}{c|c|c|c|c|c|c||c|c}
					\hline
					\multirow{3}*{Dataset} & \multirow{3}*{scale} & \multicolumn{7}{c}{Methods} \\
					\cline{3-9}
					& & NLSA & IPT & SwinIR & FPformer\tnote{$\dagger$} & FPAformer\tnote{$\dagger$} & FPformer\tnote{*} & FPAformer\tnote{*} \\ 
					& & \cite{mei2021nlsa} & \cite{chen2021ipt} & \cite{liang2021swinir} & (ours) & (ours)  & (ours) & (ours) \\
					\hline
					Set5 & 2 & 38.34/0.9618 & 38.37/- & \textbf{38.42/0.9623} & \textbf{38.42}/ \underline{0.9620} &  \underline{38.39}/0.9618 & 38.30/0.9616 & 38.28/0.9614 \\ 
					
					\cite{bevilacqua2012set5}& 3 & 34.85/0.9306 & 34.81/- & \underline{34.97}/\textbf{0.9318} & \underline{34.97/0.9317} & \textbf{34.98}/0.9314 & 34.92/0.9311 & 34.89/0.9308	\\
					
					& 4 & 32.59/0.9000 & 32.64/- & \textbf{32.92/0.9044} & \textbf{32.92}/\underline{0.9033} & \textbf{32.92}/0.9031 & 32.78/0.9024 & 32.82/0.9017\\  
					
					\hline
					\hline
					
					Set14 & 2 & 34.08/0.9231 & 34.43/- & \underline{34.46/0.9250} & 34.43/0.9241 & \textbf{34.49/0.9251} & 34.08/0.9224 & 34.08/0.9224 \\
					
					\cite{zeyde2010set14} & 3 & 30.70/0.8485 & 30.85/- & 30.93/\textbf{0.8534} & \underline{30.94/0.8523} & \textbf{30.99}/0.8520 & 30.85/0.8511 & 30.81/0.8503  \\
					
					& 4 & 28.87/0.7891 & 29.01/- & \underline{29.09}/\textbf{0.7950} & \textbf{29.10}/\underline{0.7939} & 29.07/0.7928 & 29.03/0.7922 & 29.00/0.7909 \\  
					
					\hline
					\hline
					
					Manga109 & 2 & 39.59/0.9789 & -/- & 39.92/\textbf{0.9797} & \underline{39.93/0.9795} & \textbf{39.95}/\underline{0.9795} & 39.70/0.9791 & 39.66/0.9789 \\ 
					
					\cite{matsui2017manga109} & 3 & 34.57/0.9508 & -/- & 35.12/\textbf{0.9537} & \underline{35.15/0.9532} & \textbf{35.22}/\underline{0.9532} & 34.90/0.9522 & 34.96/0.9520 \\
					
					& 4 & 31.27/0.9184 & -/- & 32.03/\textbf{0.9260} & \underline{32.05/0.9251} & \textbf{32.09}/0.9247 & 31.80/0.9227 & 31.87/0.9229 \\  
					\hline
			\end{tabular}}
		\end{threeparttable}
	\end{center}
\end{table*}

\begin{table*}[bth!]
	\caption{Comparison with state-of-the-art methods for grayscale and color image denoising on benchmark datasets in terms of average PSNR. The best and second-best results are highlighted and underlined, respectively. From left to right, the first 7 methods are task-specific trained models. The last 4 methods are task-agnostic models. $\dagger$ means Fine-tuned for a specific task. $*$ means Pre-trained.}
	\label{tab:denoise}
	\begin{center}
		\begin{threeparttable}
			{
				\begin{tabular}{c|c|c|c|c|c|c|c|c||c|c|c|c}
					\hline
					\multirow{3}*{Dataset} & \multirow{3}*{$\sigma$} & \multicolumn{11}{c}{Methods} \\
					\cline{3-13}
					& & BM3D & TNRD & DnCNN & IPT & SwinIR & FPformer\tnote{$\dagger$} & FPAformer\tnote{$\dagger$} & DRUNet & ResTormer & FPformer\tnote{*} & FPAformer\tnote{*} \\ 
					& &  \cite{dabov2007bm3d} & \cite{chen2016tnrd} & \cite{zhang2017dncnn} &  \cite{chen2021ipt} & \cite{liang2021swinir} & (ours) & (ours) & \cite{zhang2021drunet} & \cite{zamir2022restormer} & (ours) & (ours)\\
					\hline
					BSD68 & 15 & 31.08 & 31.43 & 31.73  & - & \textbf{31.97} & \underline{31.95} & 31.93 & \underline{31.91} & \textbf{31.95} & \underline{31.91} & 31.89\\ 
					
					\cite{martin2001bsd68}& 25 & 28.57 & 28.95 & 29.23 & - & \textbf{29.50} & \underline{29.48} & 29.47 & \underline{29.48} & \textbf{29.51} & 29.45 & 29.43 \\
					
					& 50 & 25.60 & 26.01 & 26.23 & - & \textbf{26.58} & \underline{26.56} & 26.55 & \underline{26.59} & \textbf{26.62} & 26.52 & 26.52 \\  
					\hline
					\hline
					CBSD68 & 15 & 33.52 & - & 33.90 & - & \textbf{34.42} & \underline{34.40} & 34.37 & 34.30 & \textbf{34.39} & \underline{34.34} & 34.31\\ 
					
					\cite{martin2001bsd68}& 25 & 30.71 & - & 31.24  & - & \textbf{31.78} & \underline{31.76} & 31.74 & 31.69 & \textbf{31.78} & \underline{31.72} & 31.69\\
					
					& 50 & 27.38 & - & 27.95  & 28.39 & \textbf{28.56} & \underline{28.54} & 28.53 & \underline{28.51} & \textbf{28.59} & \underline{28.51} & 28.49\\
					\hline
					Kodak24 & 15 & 34.28 & - & 34.60 & - & \textbf{35.34} & \textbf{35.34} & \underline{35.30} & \underline{35.31} & \textbf{35.44} & 35.27 & 35.22 \\ 
					
					\cite{franzen1999koard24}& 25 & 32.15 & - & 32.14 & - & \textbf{32.89} & \underline{32.88} & 32.86 & \underline{32.89} & \textbf{33.02} & 32.83 & 32.79\\
					
					& 50 & 28.46 & - & 28.95 & 29.64 & \textbf{29.79} & \textbf{29.79} & \underline{29.77} & \underline{29.86} & \textbf{30.00} & 29.75 & 29.72 \\
					\hline
			\end{tabular}}
		\end{threeparttable}
	\end{center}
\end{table*}

\noindent\textbf{The learning rate schedule in fine-tuning.}
\begin{figure}[tbp!]
	\centering
	\includegraphics[width=0.4\textwidth]{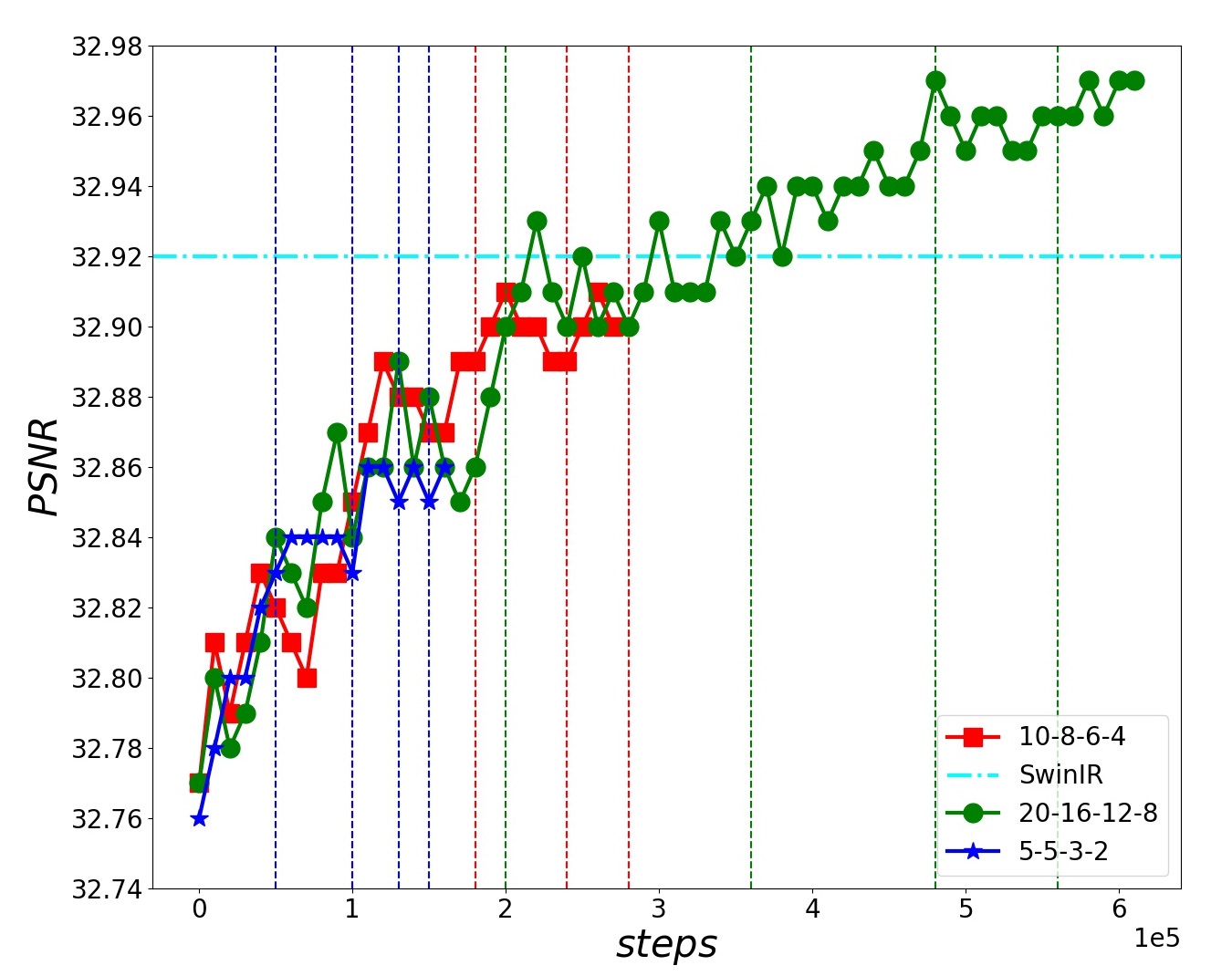}
	\caption{The learning rate schedule in fine-tuning. Fine-tuning is performed on SRx4, and results are summarized from Set5 with PSNR and SSIM.}
	\label{fig:abla_lr}
\end{figure}
After pre-training, one can further fine-tune FPformer, FPRformer, and FPAformer for a specific image restoration task. Meanwhile, fine-tuning provides an effective way to train, instead of training each task a model from scratch. We fine-tuned the pre-trained FPformer (patch size is 48) using a small learning rate, \eg 5e-5, for another 10w steps. The performance improves from 32.52dB to 32.57$\sim$32.59dB. The improvement is quite marginal. We also fine-tuned the pre-trained FPformer (patch size is 48) with a large learning rate of 2e-4 used in pre-training. The learning rate is halved at [5K, 105K, 185K, 245K, 285K] in the additional steps. The performance improves from 32.52dB to 32.70dB.

In fine-tuning the pre-trained FPformer with patch size 120, we performed three learning rate schedules starting at 2e-4. The results are summarized in Figure \ref{fig:abla_lr}. 5-5-3-2 schedule means the learning rate is halved at [5K, 55K, 105K, 135K, 155K]. It is a quick fine-tuning strategy, the improvement is about 0.1dB. In the 10-8-6-4 schedule, the learning rate is halved at [5K, 105K, 185K, 245K, 285K]. The fine-tuned models are on-par with SwinIR. In the 20-16-12-8 schedule, the learning rate is halved at [5K, 205K, 365K, 485K, 565K]. It takes about half a step for pre-training. The performance exceeds SwinIR 0.6dB to 32.98dB. It seems that pre-training provides a better initialization. Large initial learning rate and long fine-tuning benefit the performance as well. To balance performance and fine-tuning time consumption, we use a 10-8-6-4 schedule to fine-tune FPformer, FPRformer, and FPAformer. In this schedule, the fine-tuning runtime is about one-quarter of pre-training. Meanwhile, color and grayscale Gaussian denoising are joint fine-tuned. Fine-tuning these pre-trained models for 13 comparison tasks, takes $26.9\%$ time training each task-specific model from scratch. Pre-train + fine-tune training strategy provides an effective training of Vision Transformer-based image restoration models, saving energy. 

\subsection{Comparison with state-of-the-art}\label{sec:exp_sota}
In the following comparison, we choose FPformer with patch size 120. FPRformer\footnote{Due to the limited space, the performance of FPRformer is listed in the supplementary materials.} is with $T=10$ and patch size 120. FPAformer is with $T=10$, patch size 120, $ks=1$, $M=3$. The pre-trained and fine-tuned FPformer, FPRformer, and FPAformer are compared with other methods in SISR (x2, x3, and x4), color and grayscale Gaussian denoising ($\sigma$=15, 25, 50) and image JPEG deblocking ($q$=10, 20, 30, 40)\footnote{Due to the limited space, the comparison results for image JPEG deblocking are listed in the supplementary materials.}.

\noindent\textbf{Comparison with DEQ and JFB.}
We adopted and trained JFB and DEQ framework in \cite{fung2022jfb} for single image super-resolution with scale 2. The $\mathcal{F}_{\theta}$ was built on the network (38.7M parameters) for CIFAR10 in \cite{fung2022jfb}. We finetuned key hyper-parameters, \ie, $T$ and $\epsilon$. Comparison is conducted on Set5 in terms of PSNR and SSIM. JFB achieves 33.67/0.9303 (PSNR/SSIM), and DEQ achieves 38.15/0.9608. As shown in Table \ref{tab:sr}, JFB and DEQ are behind others a lot.

\noindent\textbf{SISR.}
We test comparison methods on Set5 \cite{bevilacqua2012set5}, Set14 \cite{zeyde2010set14} and Manga109 \cite{matsui2017manga109} for SISR with scale 2, 3 and 4. Following \cite{liang2021swinir}, we report PSNR and SSIM on the Y channel of the YCbCr space, as summarized in Table \ref{tab:sr}. FPformer$^\dagger$ outperforms IPT (115.5M parameters) and is on-par with SwinIR. FPAformer$^\dagger$ uses fewer model parameters and is on-par with SwinIR on Set5, outperforming SwinIR about 0.02dB on Set14 and 0.06dB on Manga109. 

\noindent\textbf{Image denoising.}
We test comparison methods on BSD68 \cite{martin2001bsd68} for grayscale denoising with noise levels 15, 25, and 50. We compare color denoising with noise levels 15, 25, and 50 on CBSD68 \cite{martin2001bsd68} and Kodak24 \cite{franzen1999koard24}. Following \cite{liang2021swinir}, we report the PSNR on the RGB channel and Y channel for color and grayscale denoising, respectively. The experimental results are summarized in Table \ref{tab:denoise}. 
In task-specific methods, the first 7 methods in Table \ref{tab:denoise}, FPformer$^\dagger$ and FPAformer$^\dagger$ outperform IPT and are on-par with SwinIR. In task-agnostic methods, the last 4 methods in Table \ref{tab:denoise}, FPformer$^*$ and FPAformer$^*$ use fewer model parameters and provide competitive performance with DRUNet (32.7M parameters) and Restormer (25.3M parameters). The number of model parameters of FPAformer is only $7.3\%$ of those in IPT,  $25.7\%$ of DRUNet, $33.2\%$ of Restormer.

\section{Conclusion}
In this work, we propose to learn the unroll of fixed-point via Transformer based models, called FPformer. By sharing parameters, we achieved a lightweight model, FPRformer. A module is proposed to analog the Anderson acceleration to boost the performance of FPRformer, called FPAformer. To fully exploit the capability of Transformer, we apply the proposed model to image restoration, using self-supervised pre-training and supervised fine-tuning. The proposed FPformer, FPRformer, and FPAformer use fewer parameters and achieve competitive performance with state-of-the-art image restoration methods and better training efficiency. FPRformer and FPAformer use only 23.21\% and 29.82\% parameters used in SwinIR models, respectively. To train these comparison models, we use only 26.9\% time used for training from scratch.

\section*{Acknowledgments}
This work was supported by the National Natural Science Foundation of China under the Grant No.61902415.

\bibliographystyle{IEEEtran}
\bibliography{egbib}

\newpage
\appendix

\section{Analysis}
In this section, we give a detailed explanation of our analysis in the main paper.

\begin{condition}\label{cond1}
	The operator $\mathcal{F}$ is contractive, that is, $\|\mathcal{F}(z')-\mathcal{F}(z)\|\leq \rho\|z'-z\|$ for some $0<\rho<1$.
\end{condition}

\begin{condition}\label{cond2}
	$(\theta_t)_{t\geq 0}$ can approximate  $\mathcal{F}$ well, that is, $\|\mathcal{F}_{\theta_t}(z)-\mathcal{F}(z)\|\leq \delta$ for some small $\delta>0$ and all $z$.
\end{condition}
Condition \ref{cond1} is the contractive property, commonly used in the optimization area. It is essential for the convergence analysis of Algorithm \ref{alg:fpf}.
While Condition \ref{cond2} is to characterize  that $(\theta_t)_{t\geq 0}$ are trained by a acceptable model. With these two conditions, we can derive the following result.
\begin{theorem}\label{th1}[Restatement of Theorem \ref{th0}]
	Assume Conditions \ref{cond1} and \ref{cond2} hold, and $(z_t)_{1\leq t\leq T}$ is generated by  \eqref{eq:train_fpformer}. Let $z^*$ be the fixed point of $\mathcal{F}$, we have
	\begin{equation} 
		\|z_T-z^*\|\leq  \rho^T\|z_0-z^*\|+\frac{\delta}{1-\rho}.
	\end{equation}
\end{theorem}

~\\

\textbf{ Proof of Theorem \ref{th1}:} Noticing that $z_{t+1}=\mathcal{F}_{\theta^t}(z_t)$, we have
\begin{align*}
	\|z_{t+1}-z^*\|&=\|\mathcal{F}_{\theta_t}(z_t)-z^*\|\\
	&=\|\mathcal{F}_{\theta_t}(z_t)-\mathcal{F}(z_t)+\mathcal{F}(z_t)-z^*\|\\
	&\leq\|\mathcal{F}_{\theta_t}(z_t)-\mathcal{F}(z_t)\|+\|\mathcal{F}(z_t)-z^*\|\\
	&\overset{\textrm{Condition}~\ref{cond2}}{\leq}\delta+\|\mathcal{F}(z_t)-z^*\|\\
	&=\delta+\|\mathcal{F}(z_t)-\mathcal{F}(z^*)\|\\
	&\overset{\textrm{Condition}~\ref{cond1}}{\leq}\delta+\rho\|z_t-z^*\|.
\end{align*}
Thus, as $T\geq 1$, we are led to
\begin{align*} 
	\|z_{T}-z^*\|&\leq \rho^t\|z_0-z^*\|+\delta\sum_{j=0}^{T-1}\rho^j\\
	&\leq\rho^T\|z_0-z^*\|+\frac{\delta}{1-\rho}.
\end{align*}

From Theorem \ref{th1}, as $T$ is large enough, it holds $\|z_T-z^*\|\approx\frac{\delta}{1-\rho}$. Noticing that $\delta$ is to reflect the level of how the trained model fit $\mathcal{F}$. Hence, the result indicates that when the $T$ is large enough and the model is trained well ($\delta>0$ is small), $z_T$ is sufficiently close to $z^*$.

Based on Theorem \ref{th1}, we can get two claims as described in the main paper. In summary, 1) as $T$ increases, the bound of $\|z_T-z^*\|$ gets small. That indicates when we use a larger $T$, the unrolling yields better results. 2) When $T=\frac{\ln\frac{1-\rho}{\delta}}{\ln\frac{1}{\rho}}$, it holds $\|z_T-z^*\|=\mathcal{O}(\frac{\delta}{1-\rho})$. That means as $T$ is fixed as some integer, the performance is only then determined by how the model fits. Following these two claims, we conclude that the performance of FPRformer may lag behind that of FPformer because of its parameter-sharing setting. Following the analysis, we design a module analog to the Anderson acceleration to explicitly enlarge the iteration times and get $\delta$ smaller.

\section{Related works}
\subsection{Fixed-Point Finding Algorithm}
The fixed-point finding in Algorithm \ref{alg:fpf}, generates a series of $\{{z_t}\}_{t=1}^{T}$ by successively applying the contraction mapping $\mathcal{F}(.)$, given a initial point $z_0$.

\begin{algorithm}[h!]
	\small
	\begin{algorithmic}[1]
		\REQUIRE $z_0$: initial point, \\ \qquad $\epsilon$: error bound, \\ \qquad $T$, maximum iteration times.
		\ENSURE $z^*$: end point satisfying either $\epsilon$ error bound or $T$ maximum iteration 
		\STATE $t \gets 0$
		\WHILE {$\| z_{t} -  \mathcal{F}(z_{t}) \|  \geq \epsilon$ and $t \leq T$}	
		\STATE $z_{t+1} \gets  \mathcal{F}(z_{t})$
		\STATE $t \gets t+1$
		\ENDWHILE
		\STATE return $z^*$
	\end{algorithmic}
	\caption{\small {Algorithm for Fixed-Point Finding}}
	\label{alg:fpf}
\end{algorithm}

In DNN-FP methods, the contraction mapping $\mathcal{F}(.)$ is parameterized by $\theta$ or a couple of $\theta_t$ as $\mathcal{F}_{\theta}$ and $\mathcal{F}_{\theta_t}$, respectively. In both forms, the iteration process in Algorithm \ref{alg:fpf} requires quite a lot of iterations, \ie, a large $T$, to reach a feasible equilibrium point $z^*$. One should also note that the choice of hyper-parameters of $\epsilon$ and $T$ is vitally important to achieving a good performance using $\mathcal{F}_{\theta}$, whose contraction property is not well-guaranteed. When applying DNN-FP methods, repeating the modern DNN a couple of times is computation-consuming and time-consuming.

\subsection{Anderson Acceleration Algorithm for Fixed-Point Finding}
Anderson acceleration (\cite{anderson1965aa,toth2015aac}, AA) is proposed to accelerate the fixed-point finding utilizing the previous $m$ states $\{z_{t-m+i}\}_{i=1}^{m}$ to estimate the next state $z_{t+1}$, as shown in Algorithm \ref{alg:fpaa}. As line 6 in Algorithm \ref{alg:fpaa}, the weights are obtained by solving a least-squares method. Then these normalized weights (whose summation is equal to 1) are used to linearly combine the previous $m-1$ states and the latest estimate from $\mathcal{F}(z_{t})$ to generate the next state $z_{t+1}$.

\begin{algorithm}[h!]
	\small
	\begin{algorithmic}[1]
		\REQUIRE $z_0$: initial point, \\ \qquad $\epsilon$: error bound, \\ \qquad $T$, maximum iteration times, \\ \qquad $m$, depth of memory, $m \geq 1$
		\ENSURE $z^*$: end point satisfying either $\epsilon$ error bound or $T$ maximum iteration 
		\STATE $t \gets 1$
		\STATE $z_1 \gets \mathcal{F}(z_{0})$
		\WHILE {$\| z_{t} -  \mathcal{F}(z_{t}) \|  \geq \epsilon$ and $t \leq T$}	
		\STATE $m_t = \min \{m, t\}$
		\STATE Given $\{ \mathcal{G}_{t-m_t+1}, \cdots, \mathcal{G}_{t} \}$, where $\mathcal{G}_{i}=\mathcal{F}(z_{i}) - z_i$
		\STATE $\left(\alpha_1, \cdots, \alpha_{m_k} \right) = \arg\min \sum_{i=1}^{m_t}  \| \alpha_i \mathcal{G}_{t-m_t+i} \|$ \\ \qquad\qquad\qquad s.t. $\sum_{i=1}^{m_t} \alpha_i = 1$
		\STATE $z_{t+1} \gets  \sum_{i=1}^{m_t} \alpha_i \mathcal{F}(z_{t-m_t+i})$
		\STATE $t \gets t+1$
		\ENDWHILE
		\STATE return $z^*$
	\end{algorithmic}
	\caption{\small {Anderson Acceleration for Fixed-Point Finding}}
	\label{alg:fpaa}
\end{algorithm}

\section{Experiments}
\subsection{Ablation Studies}
\noindent\textbf{Removing residual connection in RSTB and changing the head number of MHSA.}
For a given patch size 48, $T=10$, and $R=2$, we conducted experiments for removing residual connections in RSTB. We also investigate the influence of different head numbers (4, 8, and 12). The results are shown in Table \ref{tab:abla3}. The experimental results show that removing residual connection in RSTB and the head number of MHSA has little impact on the performance. This is different from \cite{liang2021swinir}, in which residual connection and convolution in residual connection have a significant impact on performance. We tend to believe that SwinIR trained for a specific task, the increase of parameter number will benefit the performance. However, in our FPformer, FPRformer, and FPAformer trained for a couple of image restoration tasks, the benefit is marginal.

\begin{table}[tpb]
	\caption{The influence of Removing residual connection in RSTB and the head number of MHSA. The evaluation metric is PSNR.}
	\label{tab:abla3}
	\begin{center}
		\footnotesize
		{
			\begin{tabular}{c|c|c|c}
				\hline
				no resi- & \multicolumn{3}{c}{FPRformer ($T=10$)} \\
				\cline{2-4} 
				dual & $\#head=4$ & $\#head=8$ & $\#head=12$  \\
				\hline
				32.42 & 32.43 & 32.46 & 32.46 \\
				\hline
		\end{tabular}}
	\end{center}
\end{table}

\noindent\textbf{Combination of $N_j$s.} For a given patch size 48, $T=10$ and $R=2$, we conducted experiments for different combination of $N_1$ and $N_2$. The results are summarized in Table \ref{tab:abla2}. The corresponding FPRformers achieve almost the same performance. For $R=3$, we have observed a phenomenon similar to that in the $R=2$ scenario.
\begin{table}[tpbh!]
	\caption{For a given patch size 48, $T=10$ and $R=2$, the influence of different combination of $N_1$ and $N_2$. The evaluation metric is PSNR.}
	\label{tab:abla2}
	\begin{center}
		{
			\begin{tabular}{c|c|c|c|c}
				\hline
				\multicolumn{5}{c}{FPRformer} \\
				\cline{1-5} 
				$N_1=1$ & $N_1=2$ & $N_1=3$ & $N_1=4$ & $N_1=5$  \\
				$N_2=9$ & $N_2=8$ & $N_2=7$ & $N_2=6$ & $N_2=5$ \\	
				\hline
				32.46 & 32.45 & 32.46 & 32.46 & 32.43 \\
				\hline
				\hline
				-& $N_1=6$ & $N_1=7$ & $N_1=8$ & $N_1=9$ \\
				-& $N_2=4$ & $N_2=3$ & $N_2=2$ & $N_2=1$ \\
				\hline
				-&32.42 & 32.43 & 32.44 & 32.43 \\
				\hline 	
		\end{tabular}}
	\end{center}
\end{table}

\begin{figure*}[tbph!]
	\centering
		\includegraphics[width=0.9\textwidth]{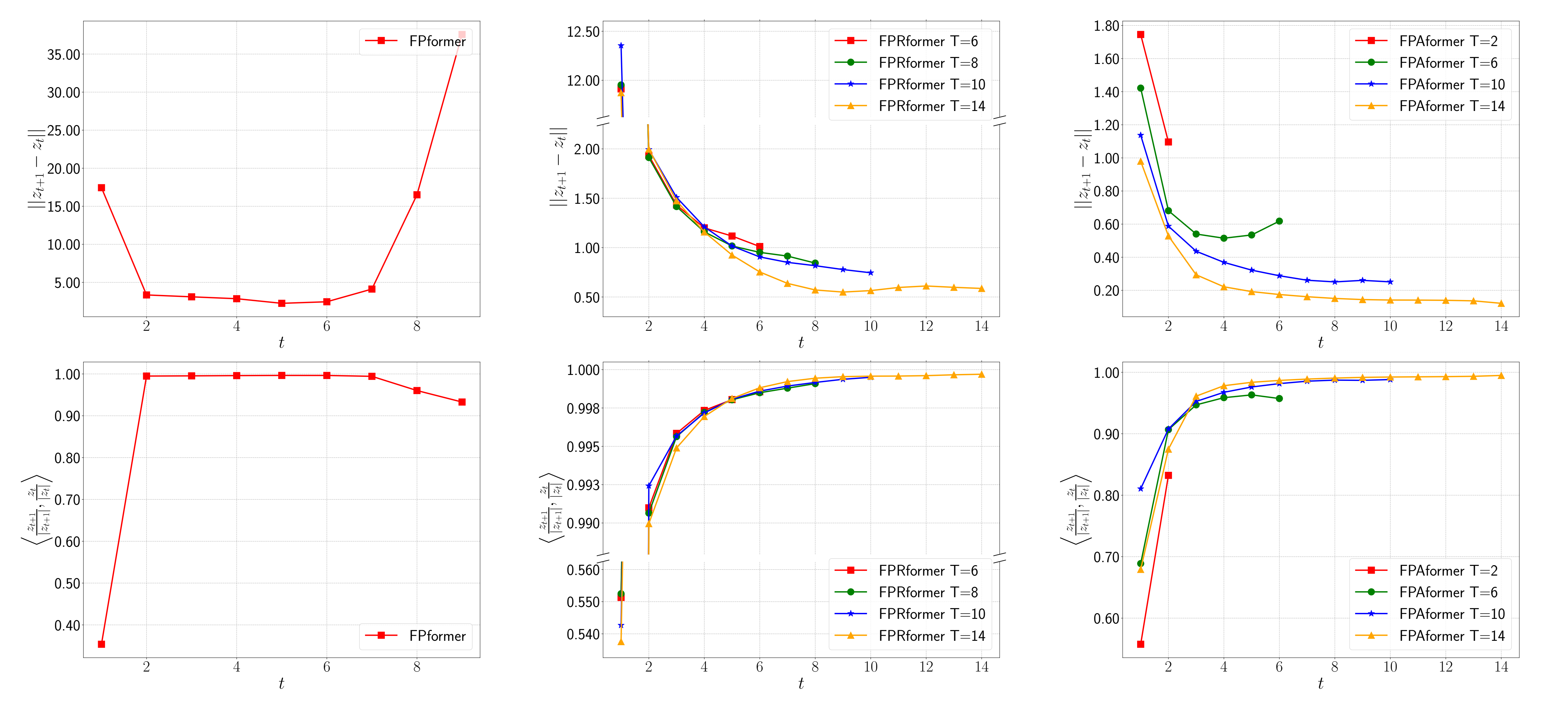}
	\caption{The $l_2$ distances and cosine similarity of input $z_t$ and output $z_{t+1}$ for each layer, for color Gaussian denoising with $\sigma=50$. From left to right, it is for FPformer, FPRformer and FPAformer.The first row is the $l_2$ distances of input $z_t$ and output $z_{t+1}$ for each layer in color Gaussian denoising with $\sigma=50$. The second row is the cosine similarity of input $z_t$ and output $z_{t+1}$ for each layer in color Gaussian denoising with $\sigma=50$.} \label{fig:abla_dist_dn}
\end{figure*}

\begin{figure*}[tbph!]
	\centering
		\includegraphics[width=0.9\textwidth]{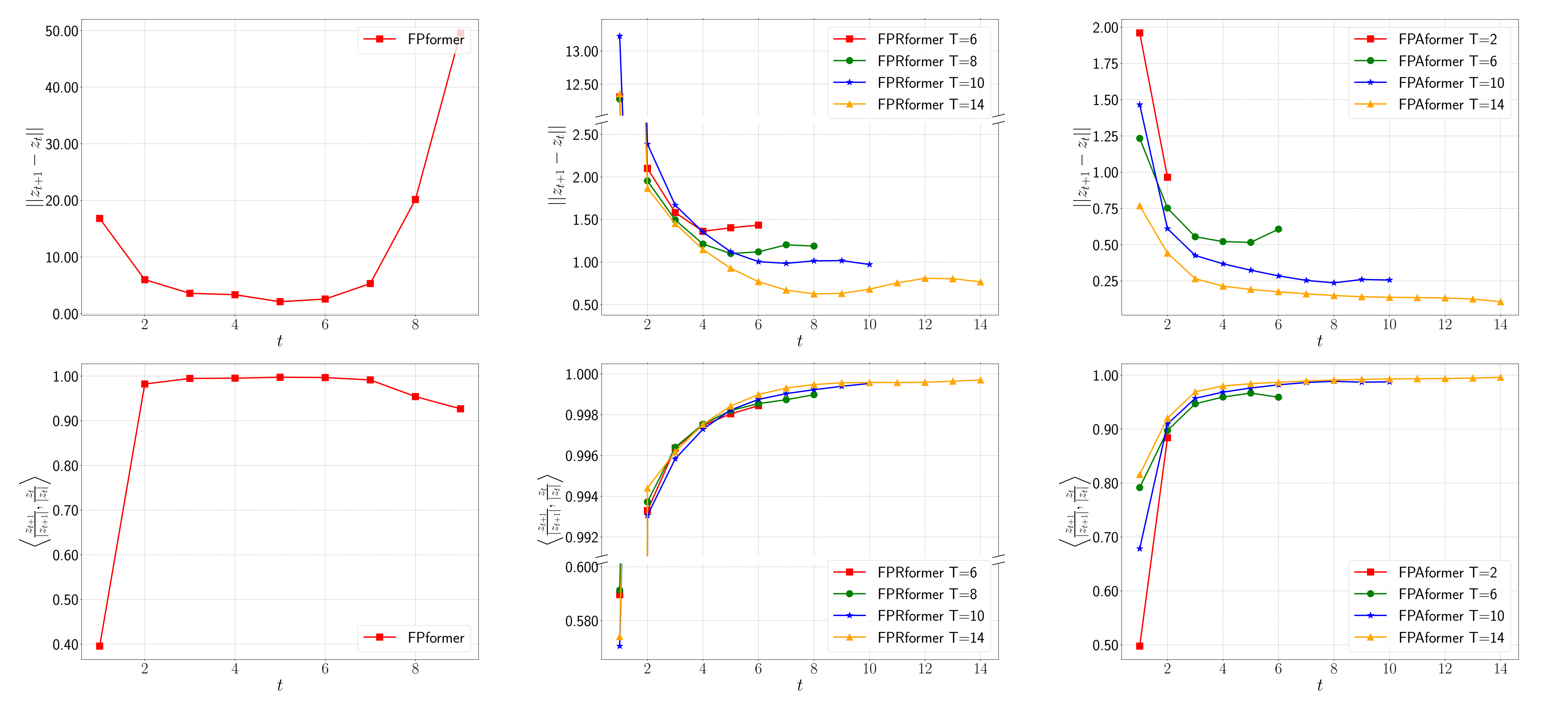}
	\caption{The $l_2$ distances and cosine similarity of input $z_t$ and output $z_{t+1}$ for each layer, for JPEG deblocking with $q=10$. From left to right, it is for FPformer, FPRformer and FPAformer. The first row is the $l_2$ distances of input $z_t$ and output $z_{t+1}$ for each layer in JPEG deblocking with $q=10$. The second row is the cosine similarity of input $z_t$ and output $z_{t+1}$ for each layer in JPEG deblocking with $q=10$.} \label{fig:abla_dist_jpeg}
\end{figure*}

\noindent\textbf{Contraction.}
The $l_2$ distances and the cosine similarity of input $z_t$ and output $z_{t+1}$ in color Gaussian denoising and image JPEG deblocking, are illustrated in Figure \ref{fig:abla_dist_dn} and \ref{fig:abla_dist_jpeg}.
The same behavior occurs in color Gaussian denoising and image JPEG deblocking as in SISR for FPformer. 
In FPRformer and FPAformer, similar trends are also presented.
For a given $T$, as $t$ increases, the $l_2$ distance is getting smaller. For $T=14$, one can also note that the l2 distance of input and output embeddings increases in the last three iterations. And this trend is quite obvious in JPEG deblocking with $q=10$ for $T=6,8,10,14$. The cosine similarity shows that these last iterations are still close to each other.

\subsection{Comparison with SOTA methods}
\noindent\textbf{SISR.}
We test comparison methods on Set5 \cite{bevilacqua2012set5}, Set14 \cite{zeyde2010set14} and Manga109 \cite{matsui2017manga109} for SISR with scale 2, 3 and 4. Following \cite{liang2021swinir}, we report PSNR and SSIM on the Y channel of the YCbCr space, as summarized in Table \ref{tab:sr_supp}. FPformer$^\dagger$ outperforms IPT (115.5M parameters) and is on-par with SwinIR. FPAformer$^\dagger$ uses fewer model parameters and is on-par with SwinIR on Set5, outperforming SwinIR about 0.02dB on Set14 and 0.06dB on Manga109. 

\begin{table*}[hbt!]
	\caption{Comparison with state-of-the-art methods for single image super-resolution on benchmark datasets in terms of average PSNR/SSIM. The best and second-best results are highlighted and underlined, respectively. $\dagger$ means Fine-tuned for a specific task. $*$ means Pre-trained.}
	\label{tab:sr_supp}
	\begin{center}
		\begin{threeparttable}
			{
				\begin{tabular}{c|c|c|c|c|c|c||c|c|c}
					\hline
					\multirow{3}*{Dataset} & \multirow{3}*{scale} & \multicolumn{8}{c}{Methods} \\
					\cline{3-10}
					& & NLSA & IPT & SwinIR & FPformer\tnote{$\dagger$} & FPAformer\tnote{$\dagger$} & FPformer\tnote{*} & FPRformer\tnote{*} & FPAformer\tnote{*} \\ 
					& & \cite{mei2021nlsa} & \cite{chen2021ipt} & \cite{liang2021swinir} & (ours) & (ours)  & (ours) & (ours) & (ours) \\
					\hline
					Set5 & 2 & 38.34/0.9618 & 38.37/- & \textbf{38.42/0.9623} & \textbf{38.42}/ \underline{0.9620} &  \underline{38.39}/0.9618 & \textbf{38.30/0.9616} & 38.26/\underline{0.9614} & \underline{38.28/0.9614} \\ 
					
					\cite{bevilacqua2012set5}& 3 & 34.85/0.9306 & 34.81/- & \underline{34.97}/\textbf{0.9318} & \underline{34.97/0.9317} & \textbf{34.98}/0.9314 & \textbf{34.92/0.9311} & 34.85/0.9306 & \underline{34.89/0.9308}	\\
					
					& 4 & 32.59/0.9000 & 32.64/- & \textbf{32.92/0.9044} & \textbf{32.92}/\underline{0.9033} & \textbf{32.92}/0.9031 & \underline{32.78}/\textbf{0.9024} & \underline{32.78/0.9017} & \textbf{32.82}/\underline{0.9017}\\  
					
					\hline
					\hline
					
					Set14 & 2 & 34.08/0.9231 & 34.43/- & \underline{34.46/0.9250} & 34.43/0.9241 & \textbf{34.49/0.9251} & \underline{34.08/0.9224} & \textbf{34.22/0.9238} & \underline{34.08/0.9224} \\
					
					\cite{zeyde2010set14} & 3 & 30.70/0.8485 & 30.85/- & 30.93/\textbf{0.8534} & \underline{30.94/0.8523} & \textbf{30.99}/0.8520 & \textbf{30.85/0.8511} & 30.74/0.8498 & \underline{30.81/0.8503} \\
					
					& 4 & 28.87/0.7891 & 29.01/- & \underline{29.09}/\textbf{0.7950} & \textbf{29.10}/\underline{0.7939} & 29.07/0.7928 & \textbf{29.03/0.7922} & \underline{29.01/0.7918} & 29.00/0.7909 \\  
					
					\hline
					\hline
					
					Manga109 & 2 & 39.59/0.9789 & -/- & 39.92/\textbf{0.9797} & \underline{39.93/0.9795} & \textbf{39.95}/\underline{0.9795} & \textbf{39.70/0.9791} & 39.61/0.9787 & 39.66/0.9789 \\ 
					
					\cite{matsui2017manga109} & 3 & 34.57/0.9508 & -/- & 35.12/\textbf{0.9537} & \underline{35.15/0.9532} & \textbf{35.22}/\underline{0.9532} & 34.90/0.9522 & 34.84/0.9517 & \textbf{34.96/0.9520} \\
					
					& 4 & 31.27/0.9184 & -/- & 32.03/\textbf{0.9260} & \underline{32.05/0.9251} & \textbf{32.09}/0.9247 & 31.80/0.9227 & 31.70/0.9221 & \textbf{31.87/0.9229} \\  
					\hline
			\end{tabular}}
		\end{threeparttable}
	\end{center}
\end{table*}

\begin{figure*}
	\centering
	\includegraphics[width=0.9\textwidth]{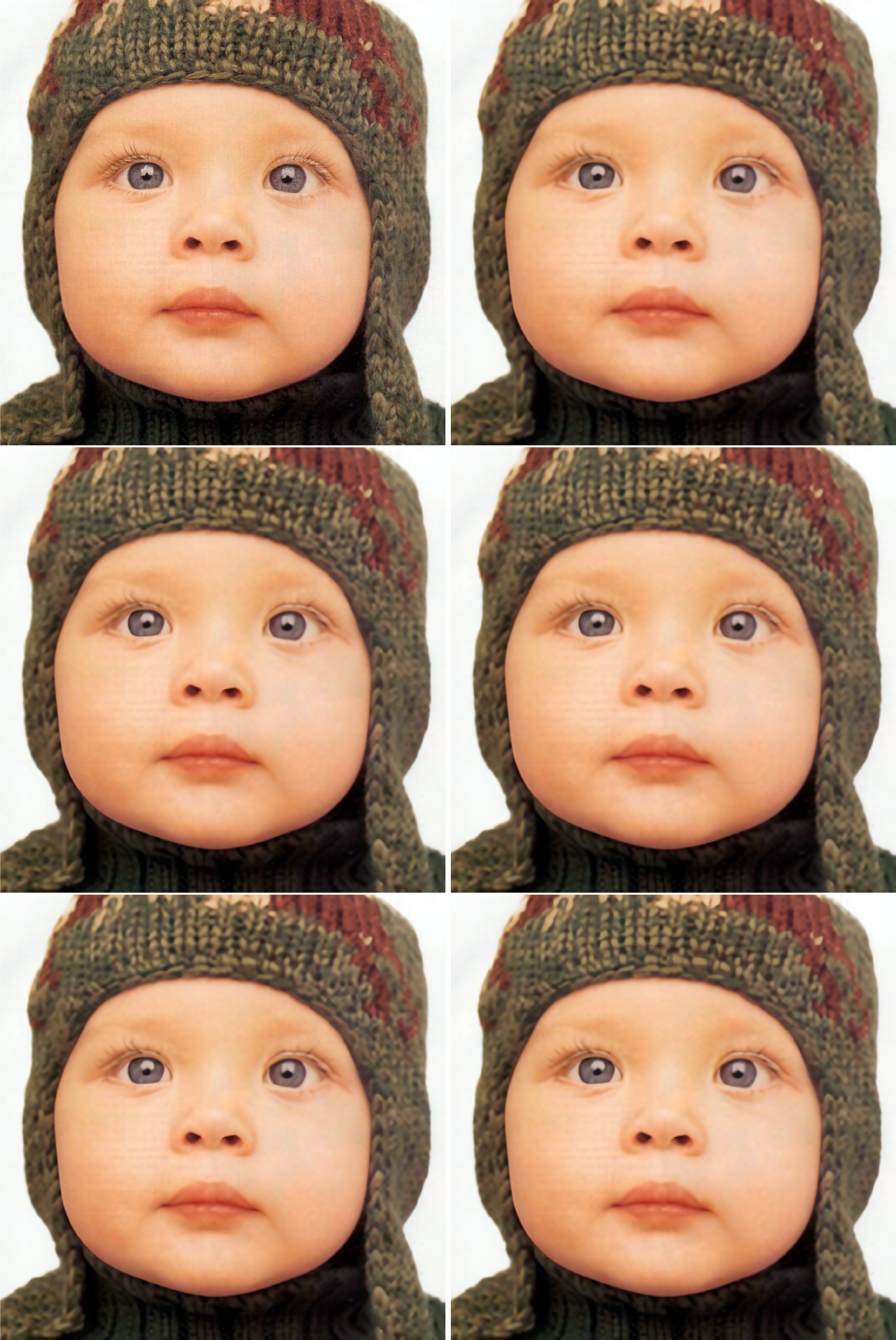}
	\caption{Visual comparison of SR×4 methods on image baby from Set5\cite{bevilacqua2012set5}. From top to bottom and left to right, the images are from original, SwinIR, FPformer{$^*$}, FPformer{$^\dagger$}, FPAformer{$^*$}, and FPAformer{$^\dagger$}. Best viewed by zooming.}
	\label{fig:sr_vis_supp}
\end{figure*}

\begin{figure*}
	\centering
	\includegraphics[width=0.9\textwidth]{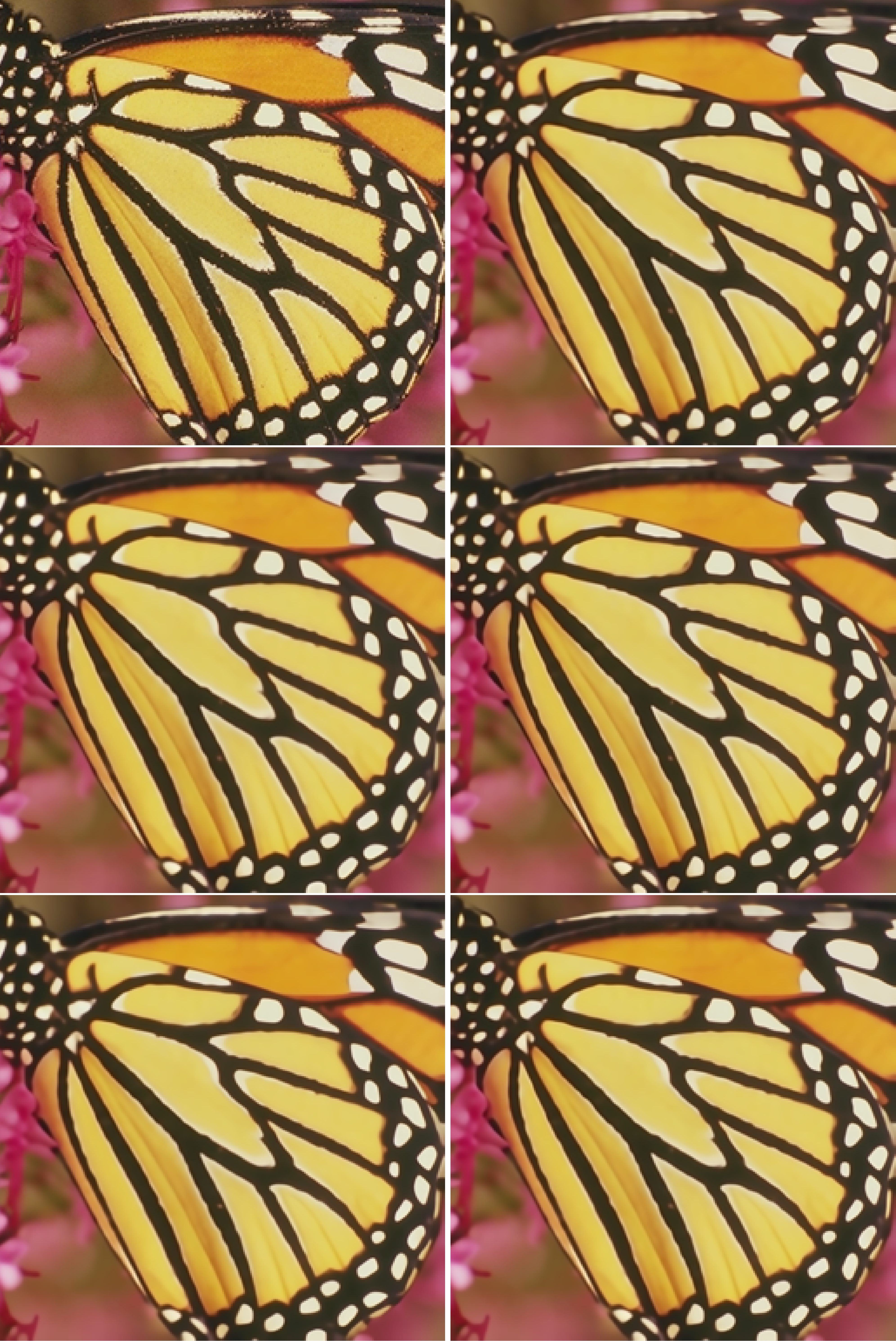}
	\caption{Visual comparison of SR×4 methods on image butterfly from Set5\cite{bevilacqua2012set5}. From top to bottom and left to right, the images are from original, SwinIR, FPformer{$^\dagger$}, FPformer{$^*$}, FPAformer{$^\dagger$}, and FPAformer{$^*$}. Best viewed by zooming.}
	\label{fig:sr_vis_supp1}
\end{figure*}

The visual comparison for SRx4 of the comparison methods are shown in Figures \ref{fig:sr_vis_supp} and \ref{fig:sr_vis_supp1}.

\noindent\textbf{Image denoising.}
We test comparison methods on BSD68 \cite{martin2001bsd68} for grayscale denoising with noise levels 15, 25, and 50. We compare color denoising with noise levels 15, 25, and 50 on CBSD68 \cite{martin2001bsd68} and Kodak24 \cite{franzen1999koard24}. Following \cite{liang2021swinir}, we report the PSNR on the RGB channel and Y channel for color and grayscale denoising, respectively. The experimental results are summarized in Table \ref{tab:denoise_supp}. 
In task-specific methods, the first 7 methods in Table \ref{tab:denoise_supp}, FPformer$^\dagger$ and FPAformer$^\dagger$ outperform IPT and are on-par with SwinIR. In task-agnostic methods, the last 5 methods in Table \ref{tab:denoise_supp}, FPformer$^*$, FPRformer$^*$, and FPAformer$^*$ use fewer model parameters and provide competitive performance with DRUNet (32.7M parameters) and Restormer (25.3M parameters). The number of model parameters of FPAformer is only $7.3\%$ of those in IPT,  $25.7\%$ of DRUNet, $33.2\%$ of Restormer.

\begin{table*}[bth!]
	\caption{Comparison with state-of-the-art methods for grayscale and color image denoising on benchmark datasets in terms of average PSNR. The best and second-best results are highlighted and underlined, respectively. From left to right, the first 7 methods are task-specific trained models. The last 5 methods are task-agnostic models. $\dagger$ means Fine-tuned for a specific task. $*$ means Pre-trained.}
	\label{tab:denoise_supp}
	\begin{center}
		\begin{threeparttable}
			{
				\begin{tabular}{c|c|c|c|c|c|c|c|c||c|c|c|c|c}
					\hline
					\multirow{3}*{Dataset} & \multirow{3}*{$\sigma$} & \multicolumn{12}{c}{Methods} \\
					\cline{3-14}
					& & BM3D & TNRD & DnCNN & IPT & SwinIR & FPformer\tnote{$\dagger$} & FPAformer\tnote{$\dagger$} & DRUNet & ResTormer & FPformer\tnote{*} & FPRformer\tnote{*} & FPAformer\tnote{*} \\ 
					& &  \cite{dabov2007bm3d} & \cite{chen2016tnrd} & \cite{zhang2017dncnn} &  \cite{chen2021ipt} & \cite{liang2021swinir} & (ours) & (ours) & \cite{zhang2021drunet} & \cite{zamir2022restormer} & (ours) & (ours) & (ours)\\
					\hline
					BSD68 & 15 & 31.08 & 31.43 & 31.73  & - & \textbf{31.97} & \underline{31.95} & 31.93 & \underline{31.91} & \textbf{31.95} & \underline{31.91} & 31.89 & 31.89\\ 
					
					\cite{martin2001bsd68}& 25 & 28.57 & 28.95 & 29.23 & - & \textbf{29.50} & \underline{29.48} & 29.47 & \underline{29.48} & \textbf{29.51} & 29.45 & 29.43 & 29.43 \\
					
					& 50 & 25.60 & 26.01 & 26.23 & - & \textbf{26.58} & \underline{26.56} & 26.55 & \underline{26.59} & \textbf{26.62} & 26.52 & 26.52 & 26.52 \\  
					\hline
					\hline
					CBSD68 & 15 & 33.52 & - & 33.90 & - & \textbf{34.42} & \underline{34.40} & 34.37 & 34.30 & \textbf{34.39} & \underline{34.34} & 34.28 & 34.31\\ 
					
					\cite{martin2001bsd68}& 25 & 30.71 & - & 31.24  & - & \textbf{31.78} & \underline{31.76} & 31.74 & 31.69 & \textbf{31.78} & \underline{31.72} & 31.66 & 31.69\\
					
					& 50 & 27.38 & - & 27.95  & 28.39 & \textbf{28.56} & \underline{28.54} & 28.53 & \underline{28.51} & \textbf{28.59} & \underline{28.51} & 28.47 & 28.49\\
					\hline
					Kodak24 & 15 & 34.28 & - & 34.60 & - & \textbf{35.34} & \textbf{35.34} & \underline{35.30} & \underline{35.31} & \textbf{35.44} & 35.27 & 35.21 & 35.22 \\ 
					
					\cite{franzen1999koard24}& 25 & 32.15 & - & 32.14 & - & \textbf{32.89} & \underline{32.88} & 32.86 & \underline{32.89} & \textbf{33.02} & 32.83 & 32.78 & 32.79\\
					
					& 50 & 28.46 & - & 28.95 & 29.64 & \textbf{29.79} & \textbf{29.79} & \underline{29.77} & \underline{29.86} & \textbf{30.00} & 29.75 & 29.70 & 29.72 \\
					\hline
			\end{tabular}}
		\end{threeparttable}
	\end{center}
\end{table*}

\begin{figure*}
	\centering
	\includegraphics[width=0.9\textwidth]{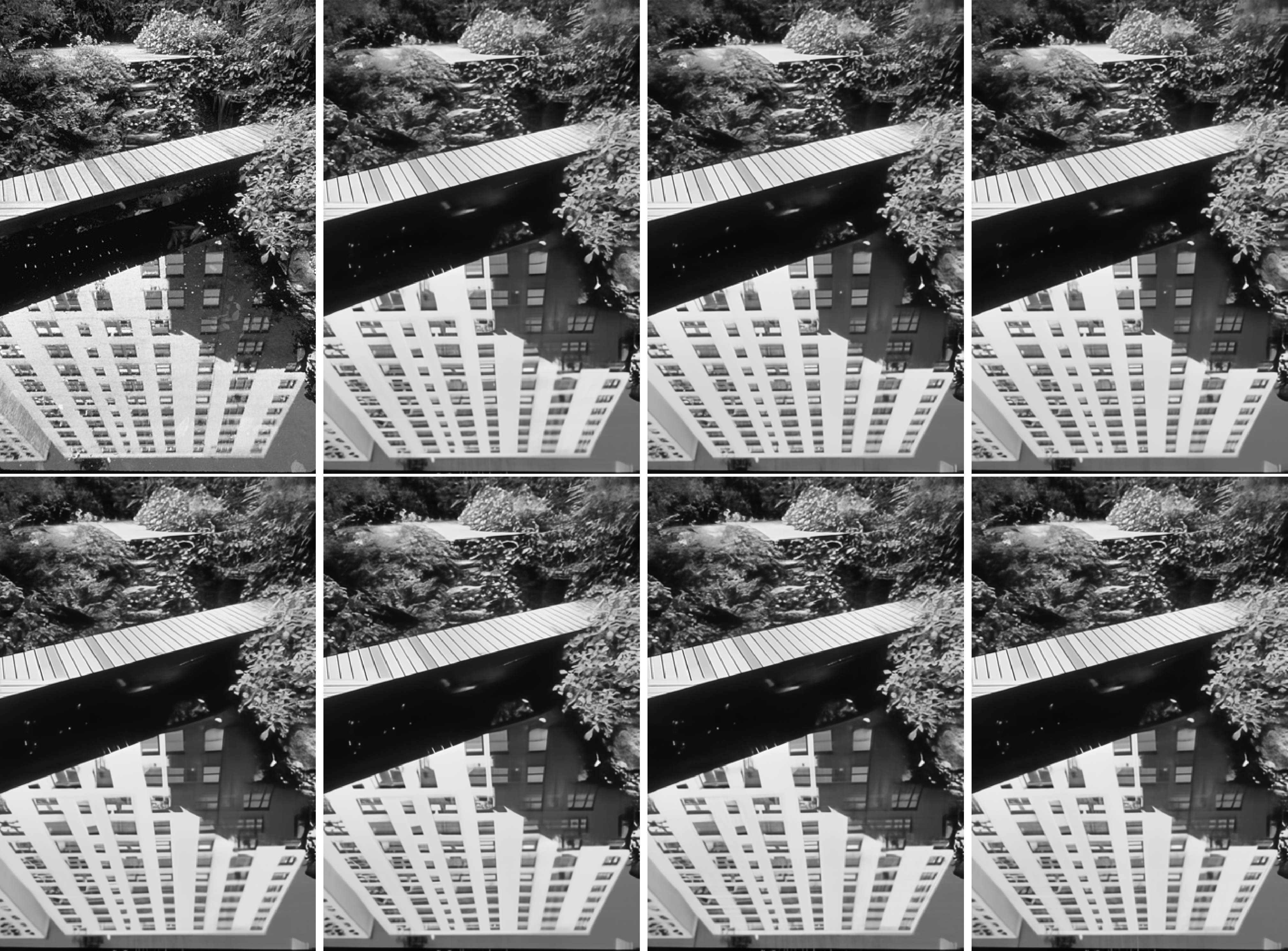}
	\caption{Visual comparison of grayscale Gaussian denoising with $\sigma=50$ methods on image test021 from BSD68\cite{martin2001bsd68}. From top to bottom and left to right, the images are from original, SwinIR, Restormer with blind setting, Restormer with nonblind setting, FPformer{$^*$}, FPformer{$^\dagger$}, FPAformer{$^*$}, and FPAformer{$^\dagger$}. Best viewed by zooming.}
	\label{fig:gdn_vis_supp}
\end{figure*}

\begin{figure*}
	\centering
	\includegraphics[width=0.9\textwidth]{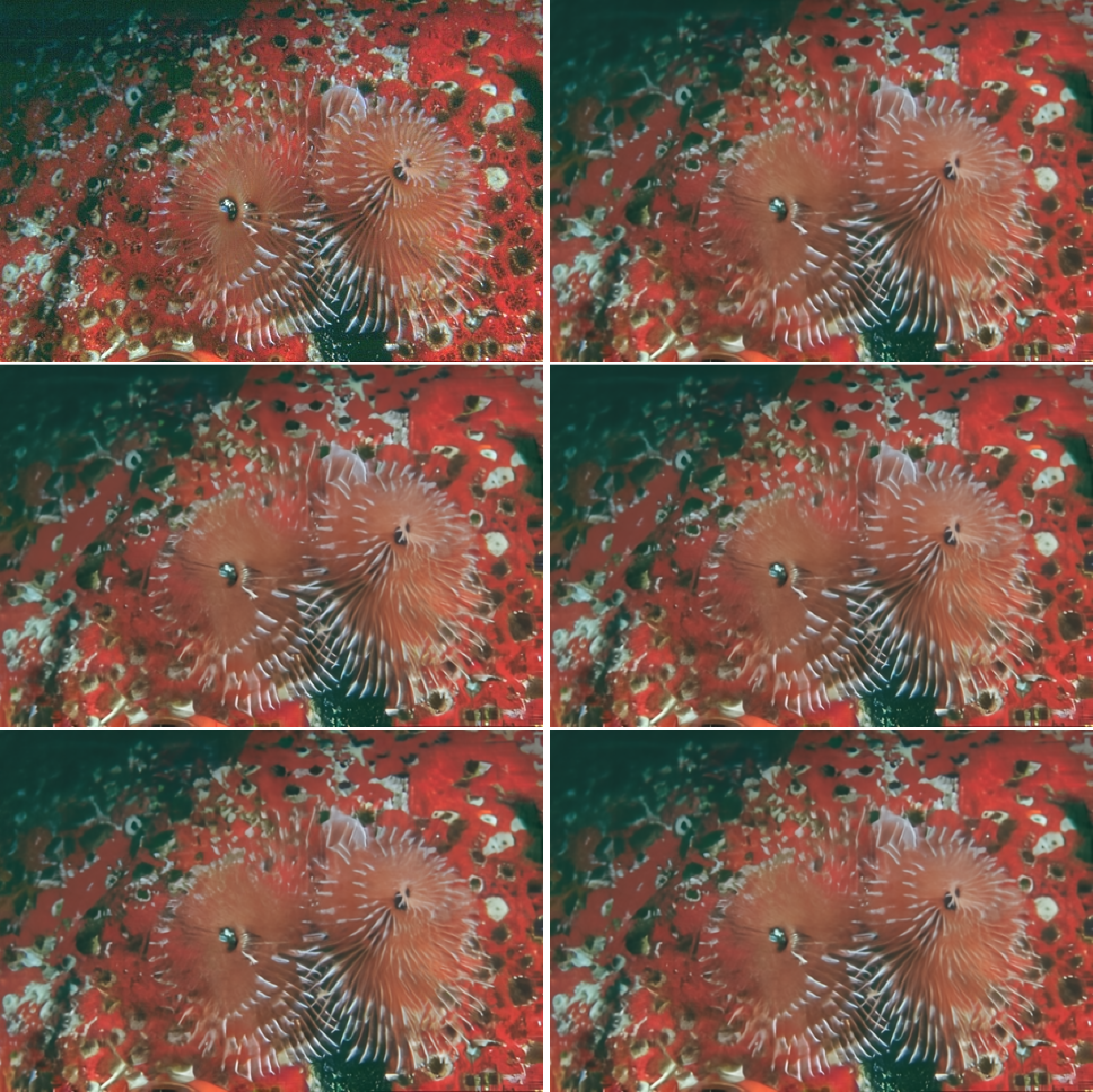}
	\caption{Visual comparison of color Gaussian denoising with $\sigma=50$ methods on image 12084 from CBSD68\cite{martin2001bsd68}. From top to bottom and left to right, the images are from original, SwinIR, FPformer{$^\dagger$}, FPformer{$^*$}, FPAformer{$^\dagger$}, and FPAformer{$^*$}. Best viewed by zooming.}
	\label{fig:cdn_vis_supp1}
\end{figure*}

The visual comparison for grayscale and color Gaussian denoising with $\sigma=50$ of the comparison methods are shown in Figures \ref{fig:gdn_vis_supp} and \ref{fig:cdn_vis_supp1}, respectively.

\noindent\textbf{Image JPEG deblocking.}
We test comparison methods on Classic5 \cite{foi2007classic5} and LIVE1 \cite{sheikh2005live1} for image JPEG deblocking with quality factor $q=10, 20, 30, 40$. Following \cite{liang2021swinir}, we report PSNR and SSIM on the Y channel, as shown in Table \ref{tab:jpeg_supp}. The FPAformer$^\dagger$ use fewer parameters, and provides very competitive performance compared with SwinIR. The FPformer$^*$ outperforms DRUNet. Both FPRformer$^*$ and FPAformer$^*$ outperform DRUNet on Classic5, and are on-par with DRUNet on LIVE1. 

\begin{table*}[tpbh!]
	\caption{Comparison with state-of-the-art methods for image JPEG deblocking on benchmark datasets in terms of average PSNR/SSIM. The best and second best results are highlighted and underlined, respectively. $\dagger$ means Fine-tuned for a specific task. $*$ means Pre-trained.}
	\label{tab:jpeg_supp}
	\begin{center}
		\begin{threeparttable}
			{
				\begin{tabular}{c|c|c|c|c|c||c|c|c}
					\hline
					\multirow{3}*{Dataset} & \multirow{3}*{$q$} & \multicolumn{7}{c}{Methods} \\
					\cline{3-9}
					& & RDN & DRUNet & SwinIR & FPAformer\tnote{$\dagger$} & FPformer\tnote{*} & FPRformer\tnote{*} & FPAformer\tnote{*} \\ 
					& & \cite{zhang2020rdn} & \cite{zhang2021drunet} & \cite{liang2021swinir} & (ours) & (ours) & (ours) & (ours) \\
					\hline
					\multirow{4}*{Classic5\cite{foi2007classic5}} & 10 & 30.00/0.8188 & 30.16/0.8234 & \textbf{30.27/0.8249} & \underline{30.23/0.8239} & \textbf{30.21/0.8235} & 30.16/0.8223 & \underline{30.17/0.8225} \\ 
					& 20 & 32.15/0.8699 & 32.39/0.8734 & \textbf{32.52/0.8748} & \underline{32.48/0.8742} & \textbf{32.43/0.8735} & 32.39/0.8729 & \underline{32.40/0.8731} \\
					& 30 & 33.43/0.8930 & 33.59/0.8949 & \textbf{33.73/0.8961} & \underline{33.71/0.8958} & \textbf{33.67/0.8952} & 33.61/0.8945 & \underline{33.63/0.8948} \\ 
					& 40 & 34.27/0.9061 & 34.41/0.9075 & \textbf{34.52/0.9082} & \underline{34.52/0.9081} & \textbf{34.49/0.9078} & 34.43/0.9073 & \underline{34.45/0.9075} \\ 
					\hline
					\hline
					\multirow{4}*{LIVE1\cite{sheikh2005live1}} & 10 & 29.67/0.8247 & 29.79/0.8278 & \textbf{29.86/0.8287} & \underline{29.83/0.8279} & \textbf{29.80/0.8276} & 29.77/0.8268 & \underline{29.78/0.8268} \\ 
					& 20 & 32.07/0.8882 & 32.17/0.8899 & \textbf{32.25/0.8909} & \underline{32.22/0.8901} & \textbf{32.20/0.8900} & 32.15/0.8895 & \underline{32.17/0.8895} \\
					& 30 & 33.51/0.9153 & 33.59/0.9166 & \textbf{33.69/0.9174} & \underline{33.65/0.9168} & \textbf{33.62/0.9166} & 33.57/0.9162 & \underline{33.59/0.9162} \\ 
					& 40 & 34.51/0.9302 & 34.58/0.9312 & \textbf{34.67/0.9317} & \underline{34.64/0.9313} & \textbf{34.61/0.9312} & 34.56/0.9308 & \underline{34.57/0.9308} \\ 
					\hline
			\end{tabular}}
		\end{threeparttable}	
	\end{center}
\end{table*}

\begin{figure*}
	\centering
	\includegraphics[width=0.9\textwidth]{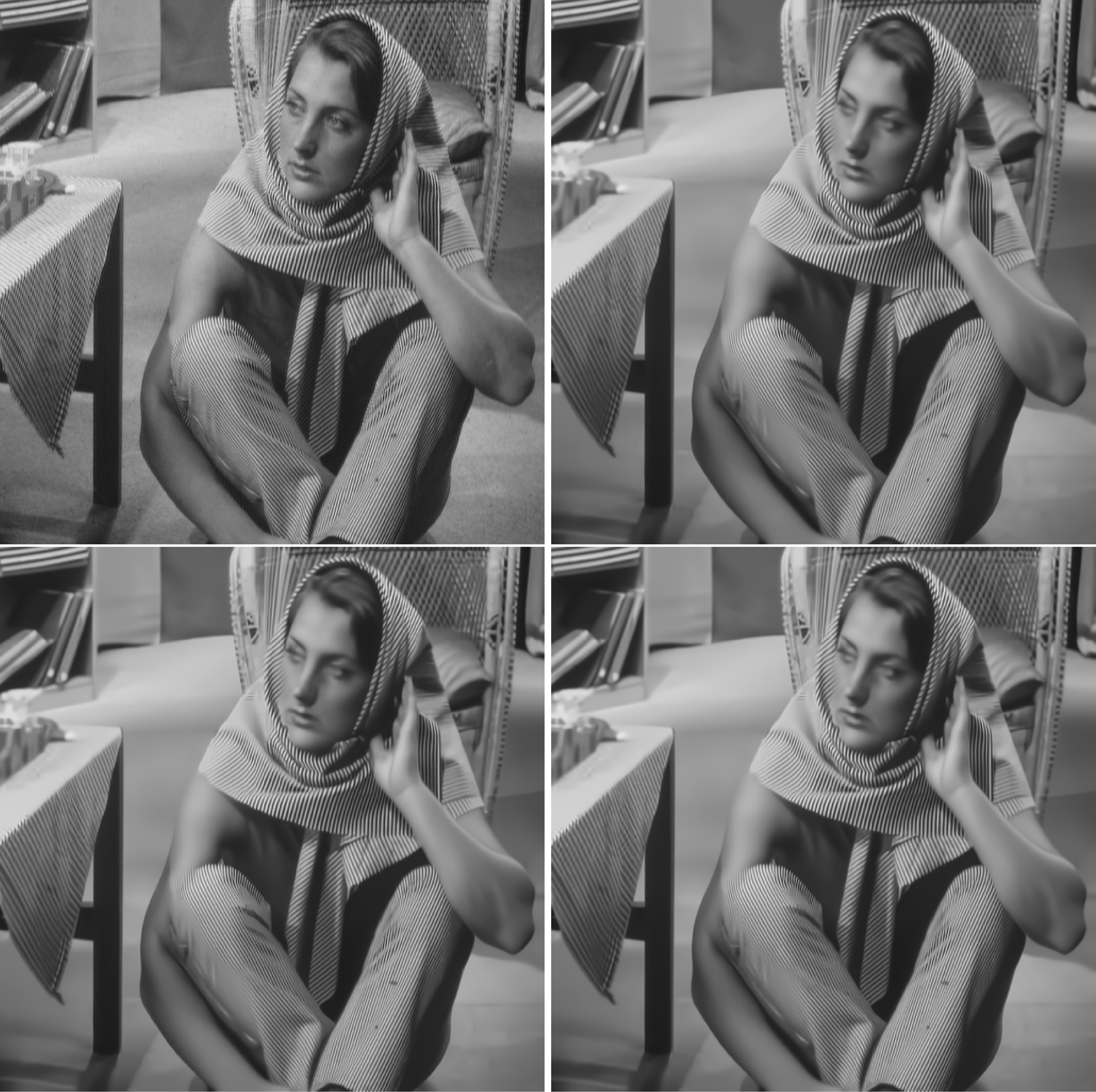}
	\caption{Visual comparison of image JPEG deblocking with quality factor $q=10$ methods on image barbara from Classic5\cite{foi2007classic5}. From top to bottom and left to right, the images are from original, SwinIR, FPAformer{$^*$}, and FPAformer{$^\dagger$}. Best viewed by zooming.}
	\label{fig:jpeg_vis_supp}
\end{figure*}

\begin{figure*}
	\centering
	\includegraphics[width=0.9\textwidth]{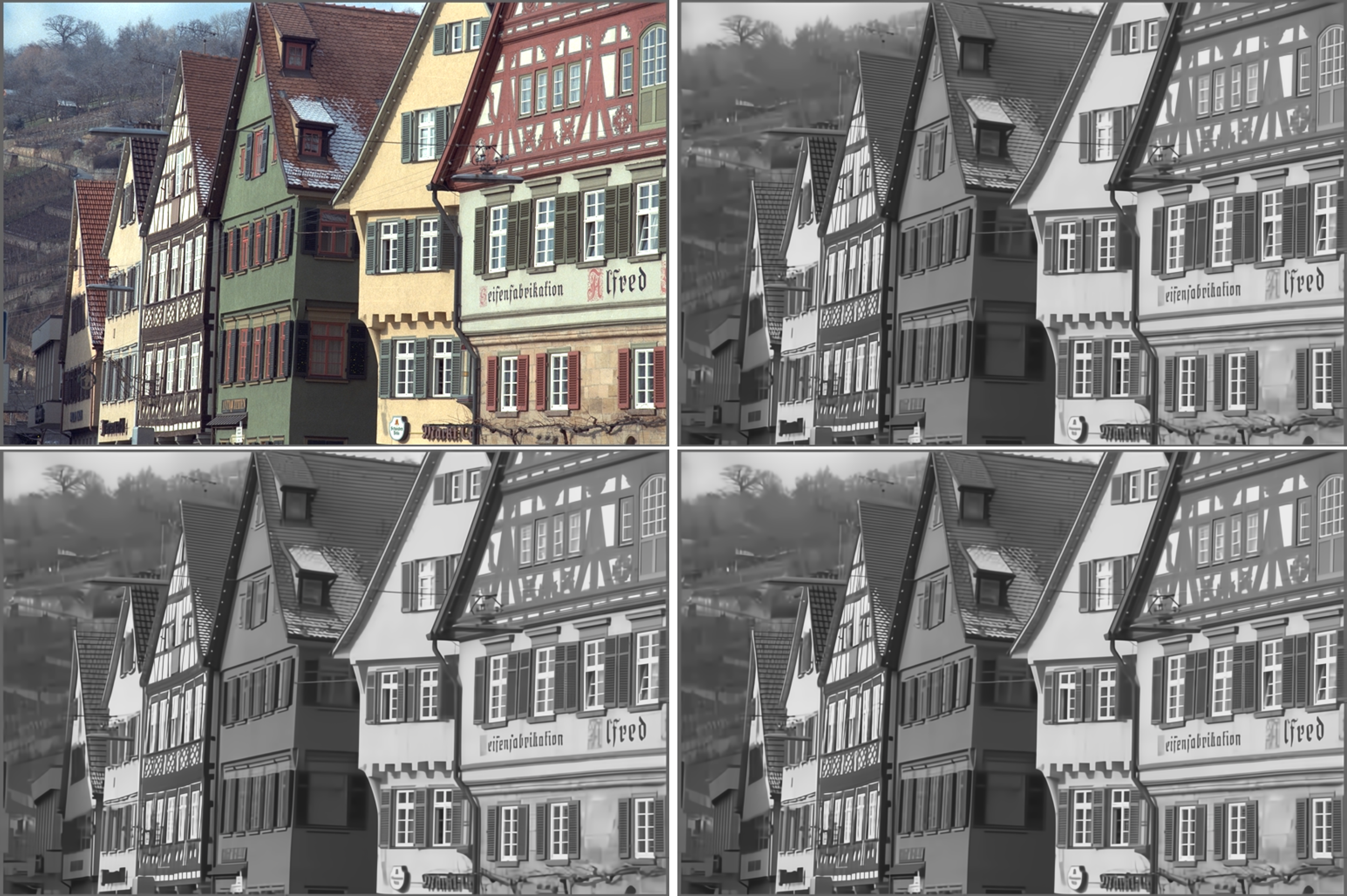}
	\caption{Visual comparison of image JPEG deblocking with quality factor $q=10$ methods on image buildings from LIVE1\cite{sheikh2005live1}. From top to bottom and left to right, the images are from original, SwinIR, FPAformer{$^\dagger$}, and FPAformer{$^*$}. Best viewed by zooming.}
	\label{fig:jpeg_vis_supp1}
\end{figure*}

The visual comparison for image JPEG deblocking with quality factor $q=10$ of the comparison methods are shown in Figures \ref{fig:jpeg_vis_supp} and \ref{fig:jpeg_vis_supp1}.

\end{document}